\RequirePackage{fix-cm}
\documentclass[pdftex,twocolumn]{svjour3}          
\smartqed  

\usepackage{bbding}
\usepackage{times}
\usepackage{epsfig}
\usepackage{graphicx}
\usepackage{caption}
\graphicspath{{figures/}}
\usepackage{amsmath}
\usepackage{amssymb}
\usepackage{autobreak}
\usepackage[colorlinks,linkcolor=red,anchorcolor=blue,citecolor=blue,CJKbookmarks=True]{hyperref}
\usepackage[utf8]{inputenc} 
\usepackage[T1]{fontenc}    
\usepackage{hyperref}       
\usepackage{booktabs}       
\usepackage{amsfonts}       
\usepackage{nicefrac}       
\usepackage{microtype}      
\usepackage{algorithm}
\usepackage{algpseudocode}
\usepackage{amsfonts,amssymb}
\usepackage{mathrsfs}
\usepackage{array}
\usepackage{bm}
\usepackage{multirow}
\usepackage{natbib}

\usepackage{times} 
\usepackage{helvet} 
\usepackage{courier} 
\usepackage{amsmath}
\usepackage{footnote}
\usepackage{multirow}
\usepackage{array}
\usepackage{algorithm}
\usepackage{algpseudocode}
\usepackage{amsfonts,amssymb}
\usepackage{booktabs}
\usepackage{xcolor}
\usepackage{verbatim}

\usepackage{ragged2e}

\usepackage{colortbl}
\usepackage{arydshln}
\usepackage{multirow}
\usepackage{multicol}
\usepackage{dashrule}
\usepackage{fancyhdr}
\usepackage{url}      
\definecolor{mygray}{gray}{.9}
\definecolor{hidden-red}{RGB}{205, 44, 36}
\definecolor{hidden-blue}{RGB}{194,232,247}
\definecolor{hidden-orange}{RGB}{243,202,120}
\definecolor{hidden-green}{RGB}{34,139,34}
\definecolor{hidden-pink}{RGB}{255,245,247}
\definecolor{hidden-black}{RGB}{20,68,106}
\definecolor{LightRed}{rgb}{1,0.92,0.92}
\definecolor{LightOrange}{rgb}{1,0.95,0.88}
\definecolor{LightYellow}{rgb}{1.0,1.0,0.84}
\definecolor{LightGreen}{rgb}{0.9,1.0,0.88}
\definecolor{LightCyan}{rgb}{0.9,1,1}
\definecolor{LightBlue}{rgb}{0.9,0.94,1}
\definecolor{LightIndigo}{rgb}{0.92,0.9,1}
\definecolor{LightMagenta}{rgb}{0.96,0.86,1}
\definecolor{DirtyWhite}{rgb}{0.96,0.96,0.96}
\definecolor{DRed}{RGB}{255,0,0}

\journalname{International Journal of Computer Vision}
\begin{document}
	
	\title{Mutually Causal Semantic Distillation Network for Zero-Shot Learning
	}
	\subtitle{}
	
	
	\author{Shiming Chen$^{1,2}$ \and Shuhuang~Chen$^{1,2}$ \and Guo-Sen~Xie$^3$ \and Xinge~You$^{1,2}$\\
		$^1$ Huazhong University of Science and Technology, Wuhan, China\\
		$^2$ National Anti-Counterfeit Engineering Research Center, Wuhan, Chna\\
		$^3$Nanjing University of Science and Technology, Nanjing, China\\
		\{gchenshiming,gsxiehm\}@gmail.com\\ 
		\{shuhuangchen,youxg\}@hust.edu.cn\\
	}
	

	

	\maketitle
	
	\begin{abstract}
		Zero-shot learning (ZSL) aims to recognize the unseen classes in the open-world guided by the side-information (\textit{e.g.}, attributes). Its key task is how to infer the latent semantic knowledge between visual and attribute features on seen classes, and thus conducting a desirable semantic knowledge transfer from seen classes to unseen ones. Prior works simply utilize unidirectional attention within a weakly-supervised manner to learn the spurious and limited latent semantic representations, which fail to effectively discover the intrinsic semantic knowledge (\textit{e.g.}, attribute semantic) between visual and attribute features. To solve the above challenges, we propose a mutually causal semantic distillation network (termed \textbf{MSDN++}) to distill the intrinsic and sufficient semantic representations for ZSL. MSDN++ consists of an attribute$\rightarrow$visual causal attention sub-net that learns attribute-based visual features, and a visual$\rightarrow$attribute causal attention sub-net that learns visual-based attribute features. The causal attentions encourages the two sub-nets to learn causal vision-attribute associations for representing reliable features with causal visual/attribute learning. With the guidance of semantic distillation loss, the two mutual attention sub-nets learn collaboratively and teach each other throughout the training process. Extensive experiments on three widely-used benchmark datasets (\textit{e.g.}, CUB, SUN, AWA2, and FLO) show that our MSDN++ yields significant improvements over the strong baselines, leading to new state-of-the-art performances.
		
	\end{abstract}
	
	\section{Introduction and Motivation}
	Visual recognition is a critical task in computer vision, which has attracted sig­nificant attention in recent years due to its numerous applications in real-world, \textit{e.g.}, autonomous driving  \citep{Hu2023PlanningorientedAD}, smart healthcare  \citep{Bai2021AdvancingCD}, and intelligent surveillance  \citep{Collins2000ASF}. Over the past decade, deep learning techniques and large-scale datasets have con­tributed to remarkable advancements in the performance of visual recognition systems  \citep{He2016DeepRL,wang2020region,wang2020suppressing}. However, existing visual models are often limited by their closed-world assumptions, where all possible classes are known in advance, and all data is given at once. Such assumptions are not applicable in practical sce­narios, where novel or previously unseen classes can arise, and data may be continually coming or decentralized due to data privacy concerns. For example, an autonomous vehicle encounters new traffic patterns, a medical AI system detects new diseases, or an intelligent system en­counters a criminal wearing a new type of disguise or clothing. Based on the prior knowledge of seen classes, humans have a remarkable ability to recognize new concepts (classes) using shared and distinct attributes of both seen and unseen classes  \citep{Lampert2009LearningTD}. 
	
	\begin{figure*}[t]
		\small
		\begin{center}
			\includegraphics[width=1\linewidth]{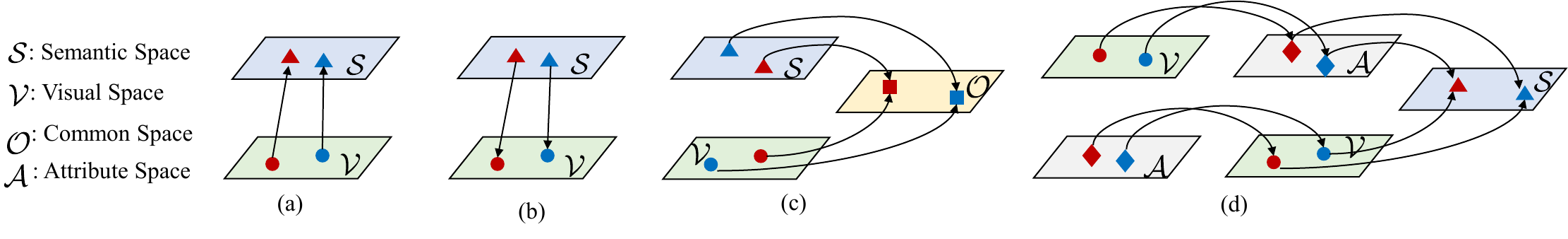}
			\\
			
			\caption{Four investigated ZSL paradigms. (a) Embedding-based method. (b) Generative method. (c) Common space learning method. (d) Ours proposed mutually causal semantic distillation network (MSDN++). The semantic space $\mathcal{S}$ is represented by the class semantic vector annotated by humans based on the attribute descriptions. The visual space $\mathcal{V}$ is learned by a network backbone (\textit{e.g.}, ResNet101  \citep{He2016DeepRL}). The common space $\mathcal{O}$ is a shared latent space between visual mapping and semantic mapping. The attribute space $\mathcal{A}$ is learned by a language model (\textit{e.g.}, Glove  \citep{Pennington2014GloveGV}). Filled triangles, circles, squares and diamonds denote the sample features in $\mathcal{S}$, $\mathcal{V}$, $\mathcal{O}$ and $\mathcal{A}$, respectively.} 
			\label{fig:paradigm}
		\end{center}
	\end{figure*}

	Analogously, zero-shot learning (ZSL) is proposed under a challenging image classification setting to mimic the human cognitive process  \citep{Larochelle2008ZerodataLO,Palatucci2009ZeroshotLW}.  ZSL aims to tackle the unseen class recognition problem by transferring knowledge from seen classes to unseen ones. It is usually based on the assumption that both seen and unseen classes can be described through the shared semantic descriptions (\textit{e.g.}, attributes)  \citep{Lampert2014AttributeBasedCF}. Based on the classes classified in the testing phase, ZSL methods can be grouped into conventional ZSL (CZSL) and generalized ZSL (GZSL)  \citep{Xian2017ZeroShotLC}, where CZSL aims to predict unseen classes, while GZSL can predict both seen and unseen ones. Since GZSL is more realistic and challenging  \citep{Pourpanah2020ARO}, some ZSL methods only focus on the GZSL setting  \citep{Liu2018GeneralizedZL,Han2020LearningTR, Han2021ContrastiveEF, Chen2021FREE,Huynh2020FineGrainedGZ}. In this work, we sufficiently evaluate our method both in the two settings.

	In the ZSL, an unseen sample shares different partial information with a set of samples of seen classes, and this partial information is represented as the abundant semantic knowledge of  attributes (\textit{e.g.}, “bill color yellow”, “leg color red” on CUB dataset)  \citep{Wang2021RegionSA,Chen2022TransZeroAT,Chen2023TransZeroCA,ChenDLLWLT24}. To this end, the key challenge of ZSL is how to infer the latent semantic knowledge between visual and attribute features on seen classes, allowing desirable knowledge transfer to unseen classes for effective visual-semantic match. Targeting on this goal, some attention-based ZSL methods  \citep{Xie2019AttentiveRE,Xie2020RegionGE,Zhu2019SemanticGuidedML,Xu2020AttributePN,Liu2021GoalOrientedGE,Chen2022GNDANGN,Chen2022TransZeroAT,Chen2023TransZeroCA,ChenHYS25} take attention mechanism to discover discriminative part/fine-grained visual features, which match the semantic representations more accurately, enabling significant ZSL performance. Unfortunately, i) they simply utilize unidirectional attention, which only focuses on limited semantic alignments between visual and attribute features without any further sequential learning; ii) {they learn the attention  with an attention modules simply supervised by the classical loss function, which only explicitly supervises the final prediction but ignores the causality between the prediction and attention. As such, prior attention-based methods can only discover the spurious and limited semantic representations between visual and attribute features, resulting in undesirable semantic knowledge transfer in the ZSL.}

	In light of the above observations, we propose a mutually {causal} semantic distillation network (termed MSDN++), to explore the intrinsic semantic knowledge between visual and attribute features for advancing ZSL, as shown in Fig. \ref{fig:paradigm} {\color{red}(d)}. The core idea is to evaluate the quality of attentions by comparing the effects of facts (\textit{i.e.}, the learned attentions) and the intervention (i.e., uncorrected attentions) on the final prediction.
	Then, we maximize the effects between the two attentions (\textit{i.e.}, effect in causal inference   \citep{Pearl_2009, Pearl_2016,LiCausal2024}) to encourage the network to learn more effective visual attentions and reduce the effects of bias of training data. Meanwhile, this {causal}  mechanism is incorporated into the mutually semantic distillation network (\textit{i.e.}, MSDN  \citep{Chen2022MSDNMS}),  enabling the  network to discover the intrinsic and more sufficient semantic knowledge for feature representations. 
	
	Specifically, MSDN++ consists of an attribute$\rightarrow$visual {causal}  attention sub-net, which learns attribute-based visual features with attribute-based/causal visual learnings, and a visual$\rightarrow$attribute {{causal}  attention sub-net, which learns visual-based attribute features via visual-based/{causal}  attribute learnings. The {causal}  attentions encourages the two sub-nets to learn causal  vision-attribute associations for representing reliable features. Meanwhile, these two mutual attention sub-nets act as a teacher-student network for guiding each other to learn collaboratively and teaching each other throughout the training process. As such, MSDN++ can explore the most matched attribute-based visual features and visual-based attribute features, enabling to effectively distill the intrinsic semantic representations for desirable knowledge transfer from seen to unseen classes. Specifically, each {causal}  attention sub-net is optimized with an attribute-based cross-entropy loss with self-calibration  \citep{Zhu2019SemanticGuidedML,Huynh2020FineGrainedGZ,Xu2020AttributePN, Liu2021GoalOrientedGE,Chen2022TransZeroAT}, attribute regression loss, and causal loss. To enable mutual learning between the two sub-nets, we further introduce a semantic distillation loss that aligns each other's class posterior probabilities. Extensive experiments well demonstrate the superiority of MSDN++ over the existing state-of-the-art methods.

		A preliminary version of this work was presented as a CVPR 2022 conference paper (termed MSDN  \citep{Chen2022MSDNMS}). In this version, we strengthen the work from three aspects: i) We further analyze the limitations of existing attention-based ZSL methods, which simply utilize unidirectional attention in a weakly-supervised manner to learn the spurious and limited latent semantic representations. As such, they fail to effectively discover the intrinsic and sufficient semantic knowledge between visual and attribute features. ii) We propose two novel sub-nets to equip MSDN  with {causal}  attentions, which enable the model to learn causal vision-attribute associations for representing reliable features. iii) We conduct substantially more experiments to demonstrate the effectiveness of the proposed methods and verify the contribution of each component.

		The main contributions of this paper are summarized as follows: 
		\begin{itemize}
			\item We propose a mutually {causal}  semantic distillation network, termed MSDN++, which distills the intrinsic and sufficient semantic representations for effective knowledge transfer of ZSL.
			\item We deploy causal  visual learning and {causal}  attribute learning to encourage MSDN++ to learn the causal vision-attribute associations for representing reliable features with good generalization.
			\item We introduce a semantic distillation loss to enable mutual learning between the attribute$\rightarrow$visual {causal}  attention sub-net and visual$\rightarrow$attribute {causal}  attention sub-net in MSDN++, encouraging them to learn attribute-based visual features and visual-based attribute features by distilling the intrinsic semantic knowledge for semantic embedding representations.
			\item We conduct extensive experiments to demonstrate that our MSDN++ achieves significant performance gains over the strong ZSL baselines on four benchmarks, \textit{i.e.}, CUB  \citep{Welinder2010CaltechUCSDB2}, SUN   \citep{Patterson2012SUNAD}, AWA2  \citep{Xian2017ZeroShotLC}, and FLO \cite{Nilsback2008AutomatedFC}.
		\end{itemize}
		
		The rest of this paper is organized as follows. Section \ref{sec2} discusses related works. The proposed MSDN++ is illustrated in Section \ref{sec3}. Experimental results and discussions are provided in Section \ref{sec4}. Finally, we draw a conclusion in Section \ref{sec5}.
		
		\begin{figure*}[t]
			\small
			\begin{center}
				\includegraphics[width=1\linewidth]{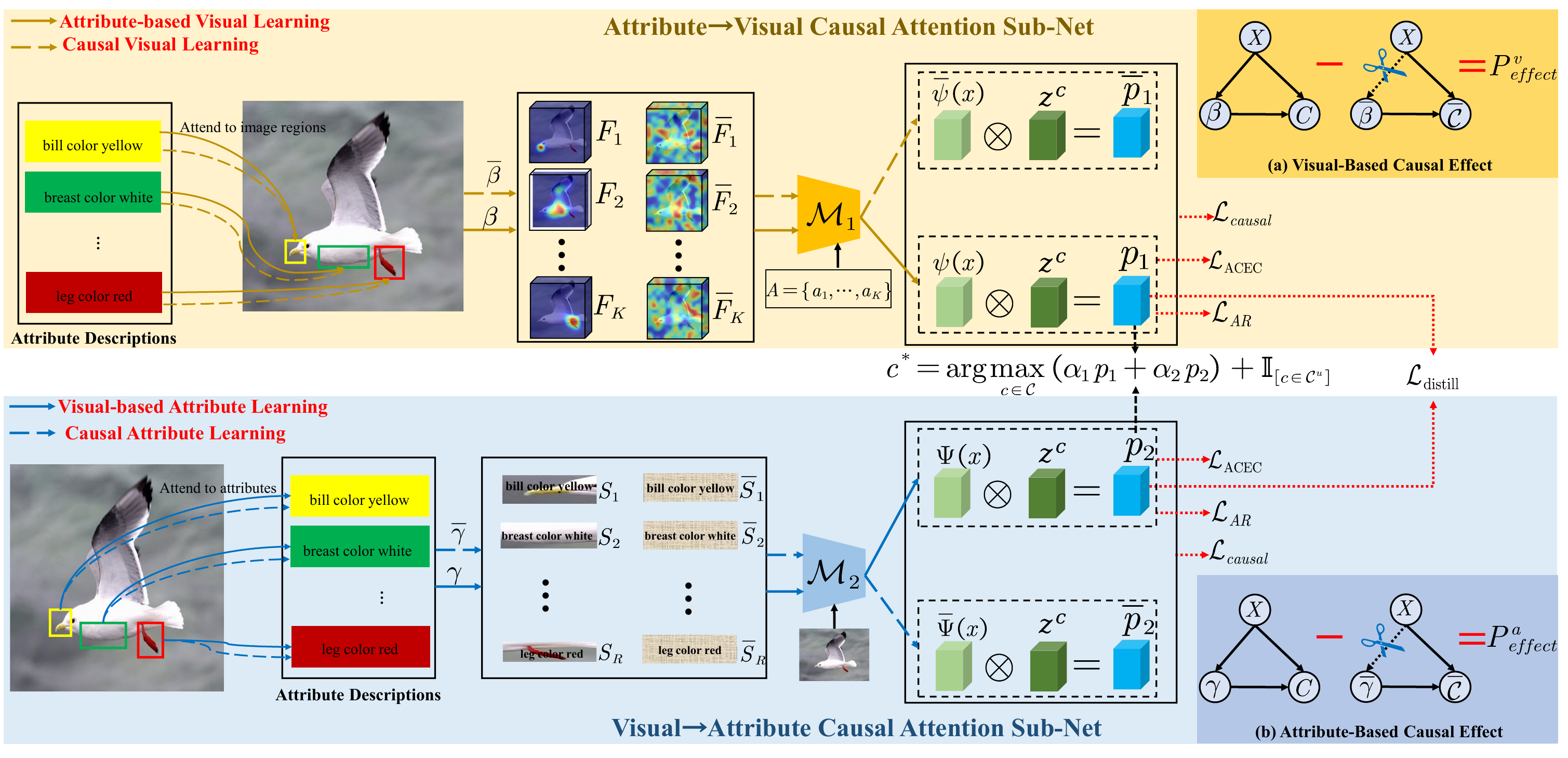}
				\\
				
				\caption{The pipeline of MSDN++. MSDN++ consists of an attribute$\rightarrow$visual {causal}  attention sub-net (AVCA) and visual$\rightarrow$attribute {causal}  attention sub-net (VACA). AVCA learns the attribute-based visual features $F$  with attribute-based visual learning and {causal}  visual learning, while VACA discovers the vision-based attribute features $S$ with vision-based attribute learning and { causal}  attribute learning.  Then, tow mapping functions $\mathcal{M}_1$ and $\mathcal{M}_2$ map the visual features and attribute features into semantic space as semantic representations $\psi(x)$ and $\Psi(x)$, respectively. A semantic distillation loss $\mathcal{L}_{distill}$ to match the probability estimates of the two sub-nets for semantic distillation (\textit{i.e.}, $p_1$ and $p_2$), enabling MSDN++ to learn intrinsic semantic knowledge. During inference, we fuse the predictions of the two sub-nets to take full use of the complementary semantic representations.} 
				\label{fig:pipeline}
			\end{center}
		\end{figure*}

		\section{Related Work}\label{sec2}
		\subsection{Zero-Shot Learning}\label{sec2.1} ZSL  \citep{Song2018TransductiveUE, Li2018DiscriminativeLO, Xian2018FeatureGN, Xian2019FVAEGAND2AF, Yu2020EpisodeBasedPG, Min2020DomainAwareVB, Han2021ContrastiveEF, Chen2021FREE, Chou2021AdaptiveAG} transfer semantic knowledge from seen classes to unseen ones by learning a mapping between the visual and attribute/semantic domains. There are several ZSL approaches targeting this goal, \textit{i.e.}, embedding-based methods, generative methods, and common space learning-based methods. As shown in Fig. \ref{fig:paradigm} {\color{red}(a)}, embedding-based methods learn a visual$\rightarrow$semantic mapping for visual-semantic interaction by mapping the visual features into the semantic space  \citep{RomeraParedes2015AnES, Akata2016LabelEmbeddingFI, Chen2018ZeroShotVR, Xie2019AttentiveRE, Xu2020AttributePN}. { Generative ZSL methods have been introduced to learn a semantic$\rightarrow$visual mapping to generate visual features of unseen classes}  \citep{Arora2018GeneralizedZL,Schnfeld2019GeneralizedZA,Xian2018FeatureGN,Li2019LeveragingTI,Yu2020EpisodeBasedPG,Shen2020InvertibleZR,Vyas2020LeveragingSA, Narayan2020LatentEF,Chen2021FREE,Chen2023EvolvingSP} for data augmentation, shown in Fig. \ref{fig:paradigm} {\color{red}(b)}. Generative ZSL methods usually base on variational autoencoders (VAEs)  \citep{Arora2018GeneralizedZL,Schnfeld2019GeneralizedZA}, generative adversarial nets (GANs)  \citep{Xian2018FeatureGN,Li2019RethinkingZL,Yu2020EpisodeBasedPG,Keshari2020GeneralizedZL,Vyas2020LeveragingSA,Chen2021FREE,ChenHYS25}, or generative flows  \citep{Shen2020InvertibleZR}. As shown in Fig. \ref{fig:paradigm} {\color{red}(c)}, common space learning is also employed to learn a common representation space for interaction between visual and semantic domains  \citep{Frome2013DeViSEAD,Tsai2017LearningRV,Schnfeld2019GeneralizedZA, Chen2021HSVA}. However, these methods still usually perform relatively sub-optimal results, since they cannot capture the subtle differences between seen and unseen classes. As such, attention-based ZSL methods  \citep{Xie2019AttentiveRE,Xie2020RegionGE,Zhu2019SemanticGuidedML,Xu2020AttributePN,Liu2021GoalOrientedGE,Chen2022GNDANGN,Chen2022TransZeroAT} utilize attribute descriptions as guidance to discover the more discriminative fine-grained features. 
		
		However, they simply utilize unidirectional attention in a weakly-supervised manner, which only focuses on spurious and limited semantic alignments between visual and attribute features without any further sequential learning. As such, the existing method fails to explore the intrinsic and sufficient semantic representations between visual and attribute features for semantic knowledge transfer of ZSL. To tackle this challenge, we will introduce a novel ZSL method based on the mutually semantic distillation learning and { causal}  attention mechanism.
		
		\subsection{Knowledge Distillation}\label{sec2.2}  To compress knowledge from a large teacher network to a small student network, knowledge distillation was proposed  \citep{Hinton2015DistillingTK}. Recently, knowledge distillation has been extended to optimize small deep networks starting with a powerful teacher network  \citep{Romero2015FitNetsHF,Parisotto2016ActorMimicDM}. By mimicking the teacher's class probabilities and/or feature representation, distilling models convey additional information beyond the conventional supervised learning target  \citep{Zhang2018DeepML, Zhai2020MultipleEB}. Motivated by these, we design a mutually semantic distillation network to learn the intrinsic semantic knowledge by semantically distilling intrinsic knowledge. The mutually semantic distillation network consists of attribute$\rightarrow$visual attention and  visual$\rightarrow$attribute attention sub-nets, which act as a teacher-student network to learn collaboratively and teach each other.
		
		\subsection{{ Causal} Inference in Vision}
		{ Causality typically includes two inferences, \textit{i.e.}, intervention inference  and counterfactual inference  \citep{Pearl_2009,Pearl_2016}, and it has been successfully applied in several areas of artificial intelligence, such as explainable machine learning}  \citep{Lv2022CausalityIR}, natural language processing  \citep{WoodDoughty2018ChallengesOU}, reinforcement learning  \citep{Kallus2018ConfoundingRobustPI} and adversarial learning  \citep{Zhang2022CausalAdvAR}. Since causal representation can alleviate the effects of dataset bias, it served as an effective tool in vision tasks  \citep{YueZS020,Rao2021CounterfactualAL, Tang2020UnbiasedSG, Chen2021HumanTP, Wang2021CausalAF, Yang2021CausalAF,LiZQSJHZ023}. For example, Chen \textit{et al.}  \citep{Chen2021HumanTP} employed a counterfactual analysis method to alleviate the over-dependence of environment bias and highlight the trajectory clues itself. Wang \textit{et al.}  \citep{Wang2021CausalAF} introduced { causal}  intervention into the visual model to
		learn causal features that are robust in any confounding context. { Li \textit{et al.}  \citep{LiZQSJHZ023} and Yue \textit{et al.}  \citep{YueZS020} bring the causal inference into few-shot learning and obtain significant performances. } In ZSL, there are clear projection domain shifts between seen and unseen classes  \citep{FuYanwei2015TransductiveMZ, Chen2023EvolvingSP}, \textit{e.g.}, the attribute of "has tail" is different for horse (seen class) and pig (unseen class), which inevitably leads the attention model to learn the spurious associations between visual and attribute features. Accordingly, we deploy { causal}  attention via intervention inference to enable the ZSL model to learn causal features, which are robust to visual bias. Accordingly, the generalization of ZSL can be improved.
		
		\section{Mutually Causal Semantic Distillation Network}\label{sec3}

		\noindent\textbf{Notation}. Assume that the training data $\mathcal{D}^{s}=\left\{\left(x_{i}^{s}, y_{i}^{s}\right)\right\}$ has $C^s$ seen classes, where $x_i^s \in  { \mathcal{X}^s}$ denotes the visual sample $i$, and $y_i^s \in \mathcal{Y}^s$ is the corresponding class label. Another set of unseen classes $\mathcal{D}^{u}=\left\{\left(x_{i}^{u}, y_{i}^{u}\right)\right\}$ has $C^u$ classes, where $x_{i}^{u}\in { \mathcal{X}^u}$ are the unseen class samples, and $y_{i}^{u} \in \mathcal{Y}^u$ are the corresponding labels. { A set of class semantic vectors/prototypes (semantic value annotated by humans according to attributes) of the class $c \in \mathcal{C}^{s} \cup \mathcal{C}^{u} = \mathcal{C}$ with ${ K}$ attributes $z^{c}=\left[z_{1}^{c}, \ldots, { z_K^{c}}\right]^{\top}= \phi(y)$ (where  $\phi(\cdot)$ is a mapping function to bridge the $z^c$ and $y$) support knowledge transfer from seen classes to unseen ones.  Here, the semantic attribute vectors of each attribute $ A=\{a_{1}, \ldots, a_{K}\}$ learned by GloVe  \citep{Pennington2014GloveGV}, it essentially is the attribute features.}
		

		{ \noindent\textbf{Overview}. As illustrated in Fig. \ref{fig:pipeline}, our MSDN++ includes an attribute $\rightarrow$visual { causal}  attention sub-net (AVCA)  and visual$\rightarrow$ attribute { causal}  attention sub-net (VACA). Under the constraint of attribute-based cross-entropy loss with self-calibration, attribute-regression loss, and causal loss, the AVCA and VACA attempt to learn attribute-based { causal}  visual features with { causal}  visual learning visual-based { causal}  attribute representations using { causal}  attribute learning, respectively. A semantic distillation loss encourages the two mutual { causal}  attention sub-nets to learn collaboratively and teach each other throughout the training process. During Inference,  we fuse the predictions of AVCA and VACA to make use the complementary semantic knowledge between the two sub-nets. }

		\subsection{Attribute$\rightarrow$Visual Causal Attention Sub-net}\label{sec3.1}
		Existing methods demonstrate that learning the fine-grained features for attribute localization is important in ZSL  \citep{Xie2019AttentiveRE, Xie2020RegionGE, Zhu2019SemanticGuidedML, Xu2020AttributePN}. In the first component of our MSDN++, we proposed an AVCA, which localizes the most relevant image regions corresponding to the attributes to extract attribute-based visual features from a given image for each attribute.  AVCA includes two streams, \textit{i.e.}, attribute-based visual learning and { causal}  visual learning. Attribute-based visual learning stream learns attribute-based visual features, which is strengthened with causality using a { causal}  visual learning stream further. 
		
		\subsubsection{Attribute-Based Visual Learning}\label{sec3.1.1}
		Attribute-based visual learning involves two inputs: a set of visual features of the image $ V=\{v_{1}, \ldots, v_{R}\}$ (where $R$ is the number of regions), such that each visual feature encodes a region in an image, and a set of semantic attribute vectors $ A=\{a_{1}, \ldots, a_{K}\}$.  AVCA attends to image regions with respect to each attribute and compares each attribute to the corresponding attended visual region features to determine the importance of each attribute. For the $k$-th attribute, its attention weight of focusing on the $r$-th region of one image is defined as:
		\begin{gather}
			\small
			\label{eq:attr_att}
			\beta_k^r = \frac{\exp \left(a_{k}^{\top} W_{1} v_{r}\right)}{\sum_{k=1}^{K} \exp \left(a_{k}^{\top} W_{1} v_{r}\right)},
		\end{gather}
		where $W_{1}$ is a learnable matrix to calculate the visual feature of each region and measure the similarity between each semantic attribute vector. As such, we get a set of attention weights $\{\beta_k^r\}_{r=1}^{R}$.
		
		AVCA then learns the attribute-based visual features for each attribute based on the attention weights. For example, AVCA obtains the $k$-th attribute-based visual feature $F_k$, which is relevant to the $k$-th attribute $a_{k}$. It is formulated as:
		\begin{gather}
			\small
			\label{eq:v_feature}
			F_k = \sum_{r=1}^{R} \beta_k^r v_r.
		\end{gather}
		Intuitively, $F_k$ captures the visual evidence to localize the corresponding semantic attribute in the image. If an image has an obvious attribute $a_k$, the model will assign a high positive score to the $k$-th attribute.  Otherwise, the model will assign a low score to the $k$-th attribute. Thus, we get a set of attribute-based visual features $F=\{F_1, F_2,\cdots, F_K\}$.
		
		After obtaining the attribute-based visual features, AVCA further maps them into the semantic embedding space using a mapping function $\mathcal{M}_1$. To encourage the mapping to be more accurate, the semantic attribute vectors $A=\{a_1, a_2,\cdots,a_K\}$ are served as support. Specifically, $\mathcal{M}_1$ matches the attribute-based visual feature $F_k$ with its corresponding semantic attribute vector $a_k$, formulated as:
		\begin{gather}
			\small
			\label{eq:m_1}
			\psi_k=\mathcal{M}_1(F_k)= a_{k}^{\top} W_2 F_k,
		\end{gather}
		where $W_2$ is an embedding matrix that maps $F$ into the semantic space. Essentially, $\psi_k$ is an attribute score that represents the confidence of having the $k$-th attribute in a given image. Finally, AVCA obtains a mapped semantic embedding $\psi(x)=\{\psi_1,\psi_2, \cdots,\psi_K\}$ for each image. 
		
		Accordingly, the final prediction can be formulated as:
		\begin{gather}
			\centering
			\label{eq:p1}
			p_1=\{\psi(x_i)\times z^1, \cdots, \psi(x_i)\times z^C\}.
		\end{gather}
		Notably, $p_1$ is the observation prediction of sample $x_i$ in AVCA.

		\subsubsection{{ Causal} Visual Learning}\label{sec3.1.2}
		There are clear projection domain shifts between seen and unseen classes in ZSL  \citep{FuYanwei2015TransductiveMZ, Chen2023EvolvingSP}, which inevitably results in the attention model to learn the spurious associations between visual and attribute features of seen and unseen classes. Accordingly, we deploy { causal}  visual learning to strengthen the attribute-based visual features with causality, enabling AVCA to learn attribute-based causal visual features.
		
		\noindent\textbf{Visual Causal Graph}.
		We reformulate the attribute-based visual learning with a visual causal graph $\mathcal{G}_v=\{\mathcal{V}_v,\mathcal{E}_v\}$. Each variable in the graph has a corresponding node in $\mathcal{V}_v$, and the causal links $\mathcal{E}_v$ describe how these variables interact with each other. As shown in Fig. \ref{fig:pipeline}(a), the nodes $\mathcal{V}_v$ in $\mathcal{G}_v$ are represented by visual features $X$, the learned attention maps $\beta$, and final prediction $\mathcal{C}$.  $(X,\beta) \rightarrow \mathcal{C}$ denotes the visual features and attention maps jointly determine the final predictions, { where we call node $X$ is the causal parent of $\beta$ and $\mathcal{C}$ is the causal child of $X$ and $\beta$.}
		
		\noindent\textbf{ Visual-Based Causal Effect}.
		Attribute-based visual learning optimizes the attention by only supervising the final prediction $\mathcal{C}$, which ignores how the learned attention maps affect the final prediction. In contrast, causal inference  \citep{Frappier2018TheBO} is a good tool to help us analyze the causalities between variables of the model. Inspired by this, we propose to adopt the causality to measure the quality of the learned attention and then improve the model by encouraging the network to produce more influential attention maps.
		Using the visual causal graph, we can analyze causalities by directly manipulating the values of attention and seeing the effect. The operation is formally called { causal}  intervention, which can be denoted as do(·). 
		
		Specifically, we take $do(\beta=\bar{\beta})$ in $\mathcal{G}_v$ to conduct { causal}  intervention, \textit{e.g.}, random attention.  	{Notably, we chose random attention as the counterfactual baseline because it represents a null intervention with minimal structural assumptions. Unlike “learned null attention” (which may still embed dataset biases) or “marginal attention” (which requires estimating a prior distribution from data), random attention is provably independent of both visual inputs and model parameters. This ensures that the intervention $do(\beta=\bar{\beta})$ truly severs the causal link between visual features and attention, fulfilling the do-operator’s requirement of an exogenous manipulation. It provides a clean, unbiased baseline to measure the true causal effect of the learned attention.} This means we replace the variable $\beta$ with the values of $\bar{\beta}$ by cutting-off the link $X\rightarrow\beta$, which without any causal associations with the visual features $X$. Following the attribute-based visual learning, we can obtain the causal visual features $\bar{F}=\{\bar{F}_1,\bar{F}_2,\\\cdots,\bar{F}_K\}$ and the causal semantic mapped embedding $\bar{\psi}(x)=\{\bar{\psi}_1,\bar{\psi}_2, \cdots,\bar{\psi}_K\}$ according to Eq. \ref{eq:v_feature} and  Eq. \ref{eq:m_1}, respectively.

		Then, the class predictions after intervention $do(\beta=\bar{\beta})$ can be obtained by:
		\begin{gather}
			\centering
			\label{eq:p_cc}
			P(do(\beta=\bar{\beta}), X=x_i)=\{\bar{\psi}(x_i)\times z^1, \cdots, \bar{\psi}(x_i)\times z^C\}.
		\end{gather}
		Accordingly, the actual effect of the learned attention on the prediction can be represented by the difference between the observed prediction $p_1=P(\beta, X=x_i)$ and its { causal}  intervention one $P(do(\beta=\bar{\beta}), X=x_i)$. It is formulated as:
		\begin{gather}
			\centering
			\label{eq:effect_1}
			P_{effect}^v(x_i)=P(\beta, X=x_i)-P(do(\beta=\bar{\beta}), X=x_i)
		\end{gather}
		{Because $\bar{\beta}$ is drawn from a uniform distribution (post softmax), it represents a maximally entropic, non-informative attention over regions. Thus, the causal effect captures how much the learned attention improves prediction compared to a completely uninformative one. We also experimented with other counterfactual distributions (e.g., uniform attention, reversed attention) in Table \ref{table:counter-att}, and results show our model is robust to the choice of $\bar{\beta}$ distribution—as long as it is independent of the input, the causal learning signal remains effective. }Therefore, we can use $P_{effect}^v(x_i)$ to evaluate the quality of the learned visual attention and improve the causality of attribute-based visual features.

		\subsection{Visual$\rightarrow$Attribute { Causal} Attention Sub-net}\label{sec3.2}
		VACA includes a visual-based attribute learning stream that learns visual-based attribute representations, and a { causal}  attribute learning that enhances the attribute features with causality. Visual-based attribute representations are complementary to the attribute-based visual features, enabling them to calibrate each other. As such, MSDN++ can discover the intrinsic semantic representations between visual and attribute features. 
		
		\subsubsection{Visual-Based Attribute Learning}\label{sec3.2.1}
		Visual-based attribute learning first attends to semantic attributes with respect to each image region. Formally, its attention weights focus on the $k$-th attribute defined as:
		\begin{gather}
			\small
			\label{eq:vis_att}
			\gamma_r^k = \frac{\exp \left(v_{r}^{\top} W_{3} a_{k}\right)}{\sum_{r=1}^{R} \exp \left(v_{r}^{\top} W_{3} a_{k}\right)},
		\end{gather}
		where $W_{3}$ is a learnable matrix, which measures the similarity between the semantic attribute vector and each visual region feature. Accordingly, VACA can get a set of attention weights $\{\gamma_r^k\}_{k=1}^{K}$, which is employed to extract visual-based attribute features, formulated as:
		\begin{gather}
			\small
			\label{eq:a_feature}
			S_r = \sum_{k=1}^{K} \gamma_r^k a_k.
		\end{gather}
		Essentially, $S_r$ is the visual-based attribute representation, which is aligned to the $F_k$. VACA further employs a mapping function $\mathcal{M}_2$ to map these visual-based attribute features $S=\{S_1,S_2,\cdots,S_R\}$ into the semantic space:
		\begin{gather}
			\small
			\label{eq:m_2}
			\hat{\Psi}_r=\mathcal{M}_2(S_r)= v_{r}^{\top} W_4 S_r,
		\end{gather}
		where $W_4$ is an embedding matrix. Given a set of $V=\{v_{1}, \ldots, v_{R}\}$, VACA obtains the mapped semantic embedding $\hat{\Psi}(x)=\{\hat{\Psi}_1,\hat{\Psi}_2, \cdots,\hat{\Psi}_R\}$ for the attributes of one image. To enable the learned semantic embedding $\hat{\Psi}(x_i)$ is $R$-$dim$ to match with the dimension of class semantic vector ($K$-$dim$), it is further mapped into semantic attribute space with $K$-$dim$, formulated as $\Psi(x_i)=\hat{\Psi}(x_i)\times Att = \hat{\Psi}(x_i) \times(V^{\top}W_{att}A)$, where $W_{att}$ is a learnable matrix. 
		
		Finally, we can get the final prediction with the mapped semantic vectors and class semantic vector:
		\begin{gather}
			\centering
			\label{eq:p2}
			p_2=\{\Psi(x_i)\times z^1, \cdots, \Psi(x_i)\times z^C\}.
		\end{gather}
		Essentially, $p_2$ is the observation prediction of sample $x_i$ in VACA.
		
		\subsubsection{{ Causal} Attribute Learning}\label{sec3.2.2}
		Similar to AVCA, VACA further improves the causality of visual-based attribute features using causal attribute learning. 
		
		\noindent\textbf{Attribute Causal Graph}.
		VACA first formulate a attribute causal graph $\mathcal{G}_a=\{\mathcal{V}_a,\mathcal{E}_a\}$. The graph consists of the corresponding nodes in $\mathcal{V}_a$, and the causal links $\mathcal{E}_a$ capturing the causal relationships between each other. As shown in Fig. \ref{fig:pipeline}(b), the nodes of $\mathcal{G}_a$ are represented by visual features $X$, the learned attribute attention maps $\gamma$, and final prediction $\mathcal{C}$.  $(X,\gamma) \rightarrow \mathcal{C}$ represents the visual features and attribute attention maps jointly determine the final predictions, where we call node $X$ is the causal parent of $\gamma$ and $\mathcal{C}$ is the causal child of $X$ and $\gamma$.
		
		\noindent\textbf{ Attribute-Based Causal Effect}.
		Based on the attribute causal graph, we can analyze causalities by directly manipulating the values of attribute attention and see the effect. For example, $do(\gamma=\bar{\gamma})$ in $\mathcal{G}_a$ conduct causal intervention, \textit{e.g.}, random attention. This means we replace the variable $\gamma$ with the values of $\bar{\gamma}$ by cutting off the link $X\rightarrow\gamma$, which does not have any causal associations with the visual features $X$. {  Following the visual-based attribute learning, we can obtain the causal attribute features $\bar{S}=\{\bar{S}_1,\bar{S}_2,\cdots,\bar{S}_R\}$, the causal mapped embedding $\bar{\Psi}(x)=\{\bar{\Psi}_1,\bar{\Psi}_2, \cdots,\bar{\Psi}_K\}$  and final prediction $\bar{p}_2$ according to Eq. \ref{eq:a_feature}, Eq. \ref{eq:m_2}, and Eq. \ref{eq:p2}, respectively.}
		
		Then, the class predictions based on the { causal intervention $do(\gamma=\bar{\gamma})$}  is formulated as:
		\begin{gather}
			\centering
			\label{eq:p_c2}
			C(do(\gamma=\bar{\gamma}), X=x_i)=\arg \max _{c \in \mathcal{C}}\bar{p}_2.
		\end{gather}
		Accordingly, the actual effect of the learned attention on the prediction is represented by the difference between the observed prediction $P(\gamma, X=x_i)$ and its { causal}  intervention one $P(do(\gamma=\bar{\gamma}), X=x_i)$. It is formulated as:
		
		\begin{gather}
			\centering
			\label{eq:effect_2}
			P_{effect}^{a}(x_i)=P(\gamma, X=x_i)-P(do(\gamma=\bar{\gamma}), X=x_i)
		\end{gather}
		To this end, the effectiveness of attribute attention can be interpreted as how the
		attention improves the final prediction compared to the wrong/intervention one. $P_{effect}^a(x_i)$ can be used to evaluate the quality of the learned attribute attention, enabling the attribute features to be more reliable.

		\begin{figure*}[t]
			\small
			\begin{center}
				\includegraphics[width=1\linewidth]{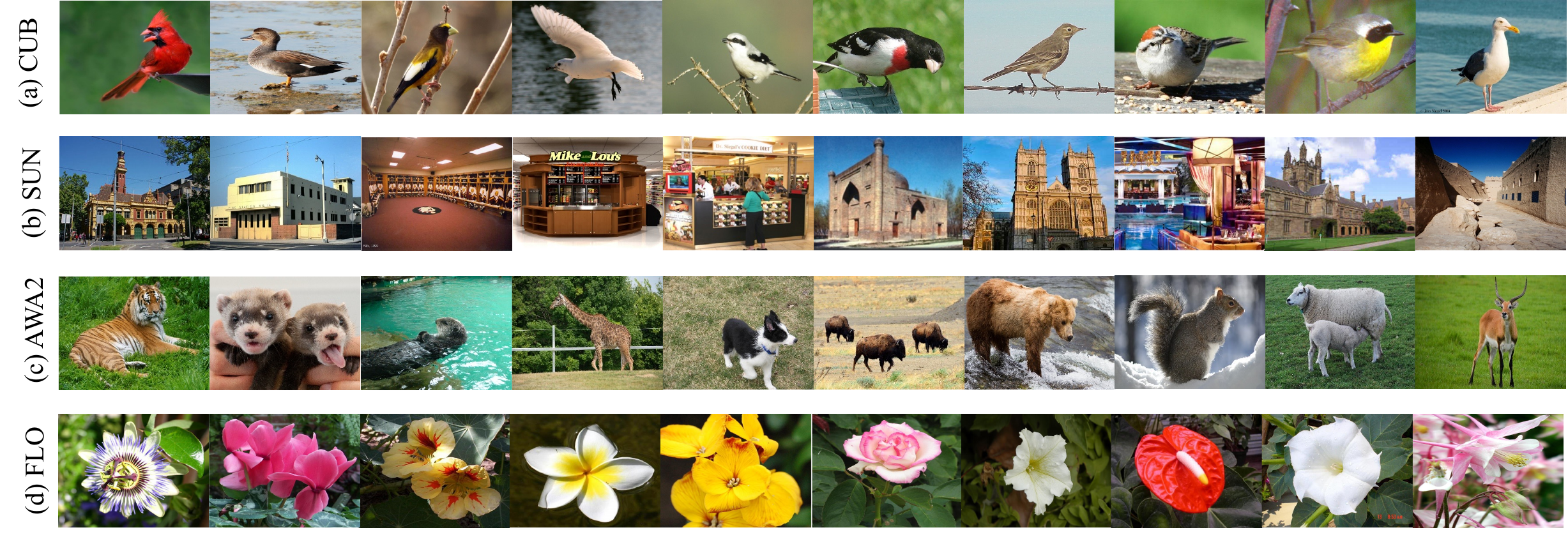}
				\caption{ Part of samples on four challenge datasets, including three fine-grained datasets (\textit{i.e.}, (a) CUB  \citep{Welinder2010CaltechUCSDB2}, (b) SUN  \citep{Patterson2012SUNAD}, and FLO \cite{Nilsback2008AutomatedFC}), and one coarse-grained dataset (\textit{i.e.}, AWA2  \citep{Xian2017ZeroShotLC}).   Each image is sampled from different classes. We find that the images of different classes in fine-grained dataset are very similar and not easy to be distinguished, while the images in the coarse-grained dataset are more easy to be recognized. For example, CUB includes similar bird images, but AWA2 consists of various animal images.} 
				\label{fig:samples}
			\end{center}
		\end{figure*}
		
		\subsection{Model Optimization}\label{sec3.3}
		To optimize { MSDN++}, each attention sub-net is trained with an attribute-based cross-entropy loss with self-calibration, attribute regression loss, and causal loss. To encourage mutual learning between the two attention sub-nets, we deploy a semantic distillation loss that aligns with each other's class posterior probabilities.

		\noindent\textbf{Attribute-Based Cross-Entropy Loss.}
		Considering the associated image and attribute embeddings are projected near their class semantic vector $z^{c}$ when an attribute is visually present in an image, we take the attribute-based cross-entropy loss with self-calibration  \citep{Zhu2019SemanticGuidedML,Huynh2020FineGrainedGZ,Xu2020AttributePN} (denoted as $\mathcal{L}_{\text{ACEC}}$) to optimize the MSDN++. $\mathcal{L}_{\text{ACEC}}$ encourages the image to have the highest compatibility score with its corresponding class semantic vector. Given a batch of $n_b$ training images $\{x_i^s\}_{i=1}^{n_b}$ with their corresponding class semantic vectors $z^c$,  $\mathcal{L}_{\text{ACEC}}$ is defined as:
		\begin{gather}
			\small
			\label{eq:L_ACEC}\
			\begin{aligned}
				\mathcal{L}_{\text{ACEC}}&=-\frac{1}{n_{b}} \sum_{i=1}^{n_{b}} [\log \frac{\exp \left(f(x_i) \times z^{c}\right)}{\sum_{\hat{c} \in \mathcal{C}^s} \exp \left(f(x_i)\times z^{\hat{c}} \right)}\\
				&-\lambda_{\text{cal}}\sum_{c^{\prime=1}}^{\mathcal{C}^u} \log \frac{\exp \left(f(x_i) \times z^{c^{\prime}} + \mathbb {I}_{\left[c^{\prime}\in\mathcal{C}^u\right]}\right)}{\sum_{\hat{c} \in \mathcal{C}} \exp \left(f(x_i) \times z^{\hat{c}} + \mathbb {I}_{\left[\hat{c}\in\mathcal{C}^u\right]}\right)}],
			\end{aligned}
		\end{gather}
		where $f(x_i)= \psi(x_i)$ for AVCA sub-net and $f(x_i)=\Psi(x_i)$ for VACA sub-net, $\mathbb {I}_{\left[c\in \mathcal{C}^u\right]}$ is an indicator function (\textit{i.e.}, it is 1 when $c\in\mathcal{C}^u$, otherwise -1), and $\lambda_{\text{cal}}$ is a weight to control the self-calibration term. Intuitively, $\mathcal{L}_{\text{ACEC}}$ encourages non-zero probabilities to be assigned to the unseen classes during training, thus MSDN++ produces a large probability for the true unseen class when given the test unseen samples.
		
		\noindent \textbf{Attribute Regression Loss.}
		We employ an attribute regression loss  \citep{Xu2022AttributePN,Chen2022TransZeroAT} to optimize MSDN++, which enables $\mathcal{M}_1$ and $\mathcal{M}_2$ to map visual/attribute features into semantic spaces close to their corresponding class semantic vectors further. Specifically, we take visual-semantic mapping as a regression problem and minimize the mean square error between the embedded attribute score $f(x_i)$ and the corresponding ground truth attribute score $z^{c}$ of a batch of $n_b$ images $\{x_i\}_{i=1}^{n_b}$:
		\begin{gather}
			\label{eq:reg-loss}
			\mathcal{L}_{\text{AR}}=\frac{1}{n_{b}} \sum_{i=1}^{n_{b}}\|f(x_i)-z^{c}\|_{2}^{2}.
		\end{gather}
		where $f(x_i)=\psi(x_i)$ and $f(x_i)=\Psi(x_i)$ in AVCV and VACA, respectively.
		
		\noindent \textbf{Causal Loss.}
		To enable MSDN++ to learn the intrinsic semantic knowledge for representing reliable features, we deploy the causal loss $\mathcal{L}_{\text{causal}}$ based on the { causal}  effect in AVCA (Eq. \ref{eq:effect_1}) and VACA (Eq. \ref{eq:effect_2}). $\mathcal{L}_{\text{causal}}$ can provide a supervision signal to explicitly guide attention learning with causal associations between visual and attribute representations, formulated as:
		\begin{gather}
			\label{eq:cau-loss}
			\begin{aligned}
				\mathcal{L}_{\text{causal}}&=\frac{1}{n_{b}} \sum_{i=1}^{n_{b}}CE(P_{effect},y_i),\\
				&=-\frac{1}{n_{b}} \sum_{i=1}^{n_{b}} \log \frac{\exp \left(f(x_i) \times z^{c}\right)}{\sum_{\hat{c} \in \mathcal{C}^s} \exp \left(f(x_i)\times z^{\hat{c}} \right)}\\
				&-\frac{1}{n_{b}} \sum_{i=1}^{n_{b}} \log \frac{\exp \left(\bar{f}(x_i) \times z^{c}\right)}{\sum_{\hat{c} \in \mathcal{C}^s} \exp \left(\bar{f}(x_i)\times z^{\hat{c}} \right)}
			\end{aligned}
		\end{gather}
		where  $P_{effect}=P_{effect}^v$ for AVCA and $P_{effect}=P_{effect}^a$ for VACA, and $CE$ is the cross-entropy.  { $f(x_i)=\psi(x_i)$ and $\bar{f}(x_i)=\bar{\psi}(x_i)$ in AVCV, while $f(x_i)=\Psi(x_i)$ and  $\bar{f}(x_i)=\bar{\Psi}(x_i)$ in  VACA.}

		\noindent \textbf{Semantic Distillation Loss.}
		We further introduce a semantic distillation loss $\mathcal{L}_{\text{distill}}$, which enables the two mutual attention sub-nets to learn collaboratively and teach each other throughout the training process. $\mathcal{L}_{\text{distill}}$ includes a Jensen-Shannon Divergence (JSD) and an $\ell_2$ distance between the predictions of the two sub-nets (\textit{i.e.}, $p_1$ and $p_2${ )}. It is formulated as:
		\begin{gather}
			\small
			\begin{aligned}
				\label{eq:L_{distill}}
				\mathcal{L}_{\text{distill}}&=\frac{1}{n_{b}} \sum_{i=1}^{n_{b}}[\underbrace{\frac{1}{2}\left(D_{K L}\left(p_1(x_i) \| p_2(x_i)\right)+D_{K L}\left(p_2(x_i) \| p_1(x_i)\right)\right)}_{\text{JSD}} \\
				&+  \underbrace{\|p_1(x_i)-p_2(x_i)\|_{2}^{2}}_{\ell_2}],
			\end{aligned}
		\end{gather}
		where
		\begin{gather}
			\small
			\label{eq:L_{distill1}}
			D_{KL}(p||q)=\sum_{c=1}^{C^s}p^c \log(\frac{p^c}{q^c}).
		\end{gather}

		\noindent\textbf{Overall Loss.} Finally, the overall loss function of MSDN++ is defined as:
		\begin{gather}
			\small
			\label{Eq:L_final}
			\mathcal{L}_{\text{total}}=  \mathcal{L}_{\text{ACEC}} + \lambda_{\text{AR}}\mathcal{L}_{\text{AR}}+ \lambda_{\text{causal}}\mathcal{L}_{\text{causal}}+ \lambda_{\text{distill}}\mathcal{L}_{\text{distill}},
		\end{gather}
		where $\lambda_{\text{AR}}$, $\lambda_{\text{causal}}$, and $\lambda_{\text{distill}}$ are weights to control the attribute regression loss, causal loss, and semantic distillation loss, respectively.

		\begin{table*}[ht]
			\small
			\centering  
			\caption{Results ~($\%$) of the state-of-the-art CZSL and GZSL models on CUB, SUN, and AWA2, including generative methods, common space-based methods, and embedding-based methods. The best and second-best results are marked in \textbf{\color{red}Red} and \textbf{ Blue}, respectively. The symbol “--” indicates no results. The symbol “{\color{red}*}” denotes attention-based methods. The symbol {\color{red}$\dagger$} denotes ZSL methods based on large-scale vision-language model.}\label{Table:SOTA}
			\resizebox{1.0\linewidth}{!}{\small
				\begin{tabular}{r|c|ccc|c|ccc|c|ccc}
					\hline
					\multirow{3}{*}{\textbf{Methods}} 
					&\multicolumn{4}{c|}{\textbf{CUB}}&\multicolumn{4}{c|}{\textbf{SUN}}&\multicolumn{4}{c}{\textbf{AWA2}}\\
					\cline{2-5}\cline{6-9}\cline{9-13}
					&\multicolumn{1}{c|}{CZSL}&\multicolumn{3}{c|}{GZSL}&\multicolumn{1}{c|}{CZSL}&\multicolumn{3}{c|}{GZSL}&\multicolumn{1}{c|}{CZSL}&\multicolumn{3}{c}{GZSL}\\
					\cline{2-5}\cline{6-9}\cline{9-13}
					\textbf{} 
					&\rm{acc}&\rm{U} & \rm{S} & \rm{H} &\rm{acc}&\rm{U} & \rm{S} & \rm{H} &\rm{acc}&\rm{U} & \rm{S} & \rm{H} \\
					
					\hline
					\rowcolor{LightRed}
					\textbf{Generative Methods} &&&&&&&&&&&&\\ 
					\rowcolor{LightRed}f-CLSWGAN~ \citep{Xian2018FeatureGN}    &57.3&43.7&57.7& 49.7&60.8&42.6&36.6&39.4&68.2&57.9&61.4&59.6\\
					\rowcolor{LightRed}f-VAEGAN-D2~ \citep{Xian2019FVAEGAND2AF}&61.0&48.4&60.1& 53.6&64.7&45.1&38.0&41.3&71.1&57.6&70.6&63.5\\
					\rowcolor{LightRed}Composer$^{\color{red}*}$~ \citep{Huynh2020CompositionalZL}&69.4&56.4&63.8&59.9&62.6& \textbf{\color{red}55.1}&22.0& 31.4&71.5& 62.1&77.3&68.8\\
					\rowcolor{LightRed}GCM-CF~ \citep{Yue2021CounterfactualZA}&--&61.0&59.7&60.3&--& 47.9&37.8& 42.2&--& 60.4&75.1&67.0\\
					\rowcolor{LightRed}FREE~ \citep{Chen2021FREE}&--&55.7&59.9&57.7&--& 47.4&37.2& 41.7&--& 60.4&75.4&67.1\\
					\rowcolor{LightRed}FREE+ESZSL~ \citep{Cetin2022CL}&--& 51.6& 60.4& 55.7&--&48.2&36.5&41.5&--&51.3&78.0& 61.8\\
					\rowcolor{LightRed}VS-Boost  \citep{Li2023VSBoostBV}&--&68.0& 68.7& 68.4&--&49.2&37.4& \textbf{\color{blue}42.5}&--&--&--&--\\
					\rowcolor{LightRed}EGG  \citep{Cavazza2023NoAT}&--& 58.6&72.3&64.7&--&--&--& --&--&50.2&87.9& 63.9\\
					\rowcolor{LightRed}ViFR \citep{ChenHYS25}& 69.1& 57.8& 62.7& 60.1& 65.6& 48.8 &35.2& 40.9& \textbf{\color{red}73.7}& 58.4& 81.4& 68.0\\
					\hline
					\rowcolor{LightOrange}\textbf{Common Space Learning} &&&&&&&&&&&&\\ 
					\rowcolor{LightOrange}DeViSE~ \citep{Frome2013DeViSEAD}&52.0&23.8&53.0&32.8&56.5&16.9&27.4&20.9&54.2&17.1&74.7&27.8\\
					\rowcolor{LightOrange}DCN~ \citep{Liu2018GeneralizedZL}&56.2&28.4&60.7&38.7&61.8&25.5&37.0&30.2&65.2&25.5&84.2&39.1\\
					\rowcolor{LightOrange}CADA-VAE~ \citep{Schnfeld2019GeneralizedZA}&59.8&51.6&53.5&52.4&61.7&47.2&35.7&40.6&63.0&55.8&75.0&63.9\\
					\rowcolor{LightOrange}SGAL~ \citep{Yu2019ZeroshotLV}  &--& 40.9 & 55.3 & 47.0 &--& 35.5 & 34.4 & 34.9& --& 52.5 & 86.3 & 65.3\\
					\rowcolor{LightOrange}CLIP$^{\color{red}\dagger}$~ \citep{Radford2021LearningTV}&--& 55.2&54.8 &55.0&-- &--&--&--&-- &--&--&--\\
					\rowcolor{LightOrange}HSVA~ \citep{Chen2021HSVA}&62.8&52.7&58.3&55.3&63.8&48.6&\textbf{\color{blue}39.0}&\textbf{\color{red}43.3}&--&59.3&76.6&66.8\\
					\rowcolor{LightOrange}CoOp$^{\color{red}\dagger}$~ \citep{Zhou2022LearningTP}&--& 49.2&63.8& 55.6&-- &--&--&--&-- &--&--&--\\
					\rowcolor{LightOrange}CoOp+SHIP$^{\color{red}\dagger}$~ \citep{Wang2023ImprovingZG}&--& 55.3& 58.9& 57.1&-- &--&--&--&-- &--&--&--\\
					\cdashline{1-13}[5pt/1pt]
					\hline 
					\rowcolor{LightBlue}\textbf{Embedding-based Methods} &&&&&&&&&&&&\\   
					\rowcolor{LightBlue}SP-AEN~~ \citep{Chen2018ZeroShotVR}      &55.4&34.7&70.6&46.6 &59.2&24.9&38.6&30.3&58.5&23.3&90.9&37.1 \\
					\rowcolor{LightBlue}SGMA$^{\color{red}*}$(NeurIPS'19)~ \citep{Zhu2019SemanticGuidedML} &71.0&36.7&71.3&48.5&--&--&--&--&68.8&37.6&87.1&52.5\\
					\rowcolor{LightBlue}AREN$^{\color{red}*}$~ \citep{Xie2019AttentiveRE}&71.8&38.9&\textbf{\color{blue}78.7}&52.1&60.6&19.0&38.8&25.5&67.9&15.6&\textbf{\color{blue}92.9}&26.7 \\
					\rowcolor{LightBlue}LFGAA$^{\color{red}*}$~ \citep{Liu2019AttributeAF}&67.6&36.2&\textbf{\color{red}80.9}&50.0&61.5&18.5&\textbf{\color{red}40.0}&25.3&68.1&27.0&\textbf{\color{red}93.4}&41.9\\
					\rowcolor{LightBlue}DAZLE$^{\color{red}*}$~ \citep{Huynh2020FineGrainedGZ}&66.0&56.7&59.6&58.1&59.4&52.3&24.3&33.2&67.9&60.3&75.7&67.1\\
					\rowcolor{LightBlue}GNDAN${\color{red}*}$~ \citep{Chen2022GNDANGN}&75.1&69.2&69.6&\textbf{\color{blue}69.4}&65.3&50.0&34.7&41.0&71.0&60.2&80.8&69.0\\
					\rowcolor{LightBlue}TransZero${\color{red}*}$~ \citep{Chen2022TransZeroAT} &\textbf{\color{blue}76.8}&\textbf{\color{blue}69.3}&68.3&68.8&65.6&\textbf{\color{blue}52.6}&33.4&40.8&70.1&61.3&82.3&70.2\\
					\rowcolor{LightBlue}APN$^{\color{red}*}$~ \citep{Xu2022AttributePN}&75.0&  67.4& 71.6& 69.4&61.5& 40.2& 35.2&37.5& 69.9& 61.9&79.4& 69.6 \\
					\rowcolor{LightBlue}{ICIS}  \citep{Christensen2023ImagefreeCI}&60.6& 45.8& 73.7& 56.5&51.8&45.2& 25.6& 32.7&64.6& 35.6 &93.3& 51.6\\ 
					\rowcolor{LightBlue}{COND+EGZSL}  \citep{Chen2024Evolutionary}&--& 45.2& 55.2& 49.6&--&--& --& --&--& 59.2&80.7& 68.3\\ 
					\rowcolor{LightBlue}{EG-part-net}  \citep{ChenDLLWLT24}&--& 64.8& 66.1& 65.4&--&--& --& --&--& \textbf{\color{blue}64.7}&78.7& \textbf{\color{blue}71.0}\\ 
					\rowcolor{LightBlue}\textbf{MSDN}$^{\color{red}*}$(\textbf{Conference Version})~ \citep{Chen2022MSDNMS}    &76.1&68.7&67.5&68.1&\textbf{\color{blue}65.8}&52.2&34.2&41.3&70.1&62.0&74.5&67.7\\
					\rowcolor{LightBlue}\textbf{MSDN++$^{\color{red}*}$ (Ours)} &\textbf{\color{red}78.5}& 	\textbf{\color{red}70.8}&	70.3 &\textbf{\color{red}70.6}&	\textbf{\color{red}67.5}&51.9&	35.4 &42.1&\textbf{\color{blue} 73.4}& \textbf{\color{red}66.5}& 79.7& \textbf{\color{red}72.5}\\	 	 
					\hline	
			\end{tabular} }
			\label{table:sota} 
		\end{table*}

		\subsection{Zero-Shot Prediction}\label{sec3.4}
		After optimization, We first obtain the embedding features of a test
		instance $x_i$ in the semantic space w.r.t. the AVCA and VACA sub-nets, \textit{i.e.}, $\psi(x)$ and $\Psi(x)$. Considering the learned semantic knowledge in the two sub-nets are complementary to each other, we fuse their predictions using two combination coefficients $(\alpha_1, \alpha_2)$ to predict the test label of $x_i$ with an explicit calibration, formulated as:
		\begin{gather}
			\small
			\label{Eq:prediction}
			c^{*}=\arg \max _{c \in \mathcal{C}^u/\mathcal{C}}(\alpha_1\psi(x_i)+\alpha_2\Psi(x_i))^{\top} \times z^{c}+\mathbb {I}_{\left[c\in\mathcal{C}^u\right]}.
		\end{gather}
		Here, $\mathcal{C}^u$/$\mathcal{C}$ corresponds to the CZSL/GZSL setting.

		\section{Experiments}\label{sec4}
		\noindent\textbf{Datasets.}  To evaluate our method, we conduct experiments on four challenging benchmark datasets, \textit{i.e.}, CUB (Caltech UCSD Birds 200)   \citep{Welinder2010CaltechUCSDB2}, SUN (SUN Attribute)  \citep{Patterson2012SUNAD}, AWA2 (Animals with Attributes 2)  \citep{Xian2017ZeroShotLC} and FLO \cite{Nilsback2008AutomatedFC}. Among them, CUB, SUN, and FLO are fine-grained datasets, whereas AWA2 is a coarse-grained dataset. Some samples are presented in Fig. \ref{fig:samples}. Following  \citep{Xian2017ZeroShotLC}, we use the same seen/unseen splits and class semantic embeddings. Specifically, CUB has 11,788 images of 200 bird classes (seen/unseen classes = 150/50) with 312 attributes. SUN consists of 14,340 images of 717 scene classes (seen/unseen classes = 645/72) with 102 attributes. AWA2 has 37,322 images of 50 animal classes (seen/unseen classes = 40/10) with 85 attributes. {FLO has  8189 images of 102 flowers classes (seen/unseen classes = 82/20) with 1024 attributes.}

		\noindent\textbf{Evaluation Protocols.} In the CZSL setting, we evaluate the top-1 accuracy on unseen classes , denoted as $acc$. In the GZSL setting, we evaluate the top-1 accuracies both on seen and unseen classes (\textit{i.e.}, $S$ and $U$). Furthermore, their harmonic mean (defined as $H =(2 \times S \times U) /(S+U)$) is also used for evaluating the performance in the GZSL setting.

		\noindent\textbf{Implementation Details.} {We take a ResNet101  \citep{He2016DeepRL} pre-trained on ImageNet as the network backbone to extract the feature map for each image without fine-tuning. Input images were resized to $448\times448$ and normalized using ImageNet statistics. The feature maps were extracted from the last convolutional layer, resulting in a fixed spatial dimension of $14\times14$ (196 regions). This consistent spatial structure was maintained throughout training and intervention. We use the RMSProp optimizer with hyperparameters (momentum = 0.9, weight decay = 0.0001) to optimize our model. We set the learning rate and batch size to 0.0001 and 50, respectively. We take random attention as the { causal}  intervention in { causal}  visual/attribute learning streams for all datasets.  Specifically, we generated random attention weights independently for each sample. For the AVCA subnet, random attention matrices of size $[K,49]$ were sampled from a Uniform(0,1) distribution and then normalized via softmax over the spatial dimension.  The same procedure was applied symmetrically to the VACA subnet. These random attentions were used only in the forward pass for causal loss computation and were detached from the gradient computation, ensuring they did not affect the learning of the actual attention modules.
		We empirically set the loss weights $\{\lambda_{\text{cal}},\lambda_{\text{AR}},\lambda_{\text{causal}},\lambda_{\text{distill}}\}$ to $\{0.05,0.03,0.3,0.001\}$, $\{0.0001, 0.01,0.0005,0.05\}$, and \\ $\{0.4,0.06,0.1,0.01\}$ for CUB, SUN, and AWA2, respectively. We set the combination coefficient $(\alpha_{1},\alpha_{2})$ to $(0.8,0.2)$, $(0.7,0.3)$, $(0.8,0.2)$ for CUB, SUN, and AWA2, respectively. }
		
		\begin{figure*}[t]
			\begin{center}
				\includegraphics[width=1.0\linewidth]{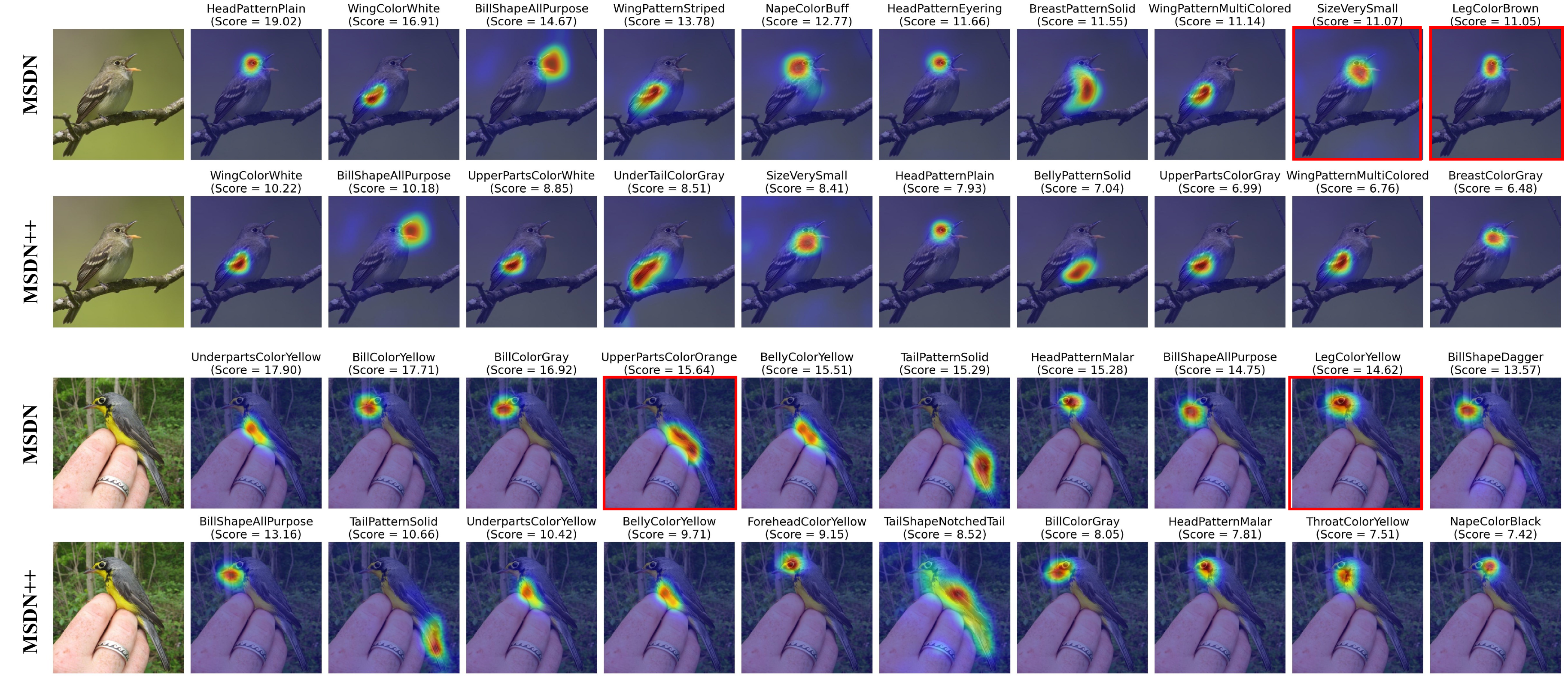}\\
				\caption{Visualization of attention maps learned by the first sub-nets of MSDN  \citep{Chen2022MSDNMS} and MSDN++ on CUB. We show the top-10 attention maps focused by models. The red boxes indicate MSDN learns the wrong attention maps that are irrelevant to the corresponding attributes.}
				\label{fig:att}
			\end{center}
		\end{figure*}
	
			\begin{table}[ht]
			\small
			\centering  
			\caption{Results ~($\%$) of the state-of-the-art  GZSL models on FLO.}
			\resizebox{1.0\linewidth}{!}{\small
				\begin{tabular}{r|ccc}
					\hline
					\multirow{3}{*}{\textbf{Methods}} 
					&\multicolumn{3}{c}{\textbf{FLO}}\\
					\cline{2-4}
					&\rm{U} & \rm{S} & \rm{H} \\
					\hline
					f-CLSWGAN~ \citep{Xian2018FeatureGN}    &59.3&74.2& 65.9\\
					f-VAEGAN-D2~ \citep{Xian2019FVAEGAND2AF}&56.8&74.9& 64.6\\
				    TF-VAEGAN~ \citep{Narayan2020LatentEF}&62.5&84.1&71.7\\
					{EG-part-net}  \citep{ChenDLLWLT24}& 61.5& 81.4& 70.1\\ 
					\textbf{MSDN}(\textbf{Conference Version})~ \citep{Chen2022MSDNMS}    &62.2&81.0&70.3\\
					\textbf{MSDN++ (Ours)}& 	69.2&	80.7 &\textbf{74.5}\\	 	 
					\hline	
			\end{tabular} }
			\label{table:sota-flo} 
		\end{table}

		\subsection{Comparision with State-of-the-Arts}\label{sec4.1}
		Our MSDN++ is an {embedding}-based method with inductive manner. To demonstrate the effectiveness and advantages of our MSDN++, we compare it with other state-of-the-art methods both in CZSL and GZSL settings, including generative methods (\textit{e.g.}, f-CLSWGAN~ \citep{Xian2018FeatureGN}, f-VAEGAN~ \citep{Xian2019FVAEGAND2AF}, Composer~ \citep{Huynh2020CompositionalZL}, E-PGN~ \citep{Yu2020EpisodeBasedPG}, TF-VAEGAN~ \citep{Narayan2020LatentEF}, IZF~ \citep{Shen2020InvertibleZR}, SDGZSL~ \citep{Chen2021SemanticsDF}, GCM-CF~ \citep{Yue2021CounterfactualZA}, FREE~ \citep{Chen2021FREE},  FREE+ESZSL~ \citep{Cetin2022CL}, VS-Boost  \citep{Li2023VSBoostBV}, EGG  \citep{Cavazza2023NoAT}, and ViFR\citep{ChenHYS25}), common space learning methods (\textit{e.g.}, DeViSE~ \citep{Frome2013DeViSEAD}, DCN~ \citep{Liu2018GeneralizedZL}, CADA-VAE~ \citep{Schnfeld2019GeneralizedZA}, SGAL~ \citep{Yu2019ZeroshotLV}, and HSVA~ \citep{Chen2021HSVA}, CLIP~ \citep{Radford2021LearningTV},  CoOp~ \citep{Zhou2022LearningTP}, CoOp+SHIP  \citep{Wang2023ImprovingZG}), and embedding-based methods (\textit{e.g.}, SP-AEN~~ \citep{Chen2018ZeroShotVR}, SGMA , AREN~ \citep{Xie2019AttentiveRE}, LFGAA~ \citep{Liu2019AttributeAF},   DAZLE~ \citep{Huynh2020FineGrainedGZ}, GNDAN~ \citep{Chen2022GNDANGN}, TransZero~ \citep{Chen2022TransZeroAT}, APN~ \citep{Xu2022AttributePN}, 
		{ ICIS}  \citep{Christensen2023ImagefreeCI}, 
		{ EG-part-net} \citep{ChenDLLWLT24},
		{ COND+EGZSL} \citep{Chen2024Evolutionary}). 
		Notably, the large-scale visual-language model based ZSL methods  \citep{Radford2021LearningTV, Zhou2022LearningTP,Wang2023ImprovingZG} can be grouped into common space learning methods intrinsically. Different to classical common space learning methods  \citep{Frome2013DeViSEAD,Schnfeld2019GeneralizedZA,Chen2021HSVA} that uses small datasets for learning a joint space of visual and semantic representations with semantic attributes, the large-scale visual-language model based ZSL methods learns the joint space with large-scale data and label prompt.
		
			\begin{figure*}[ht]
			\begin{center}
				\includegraphics[width=1.0\linewidth]{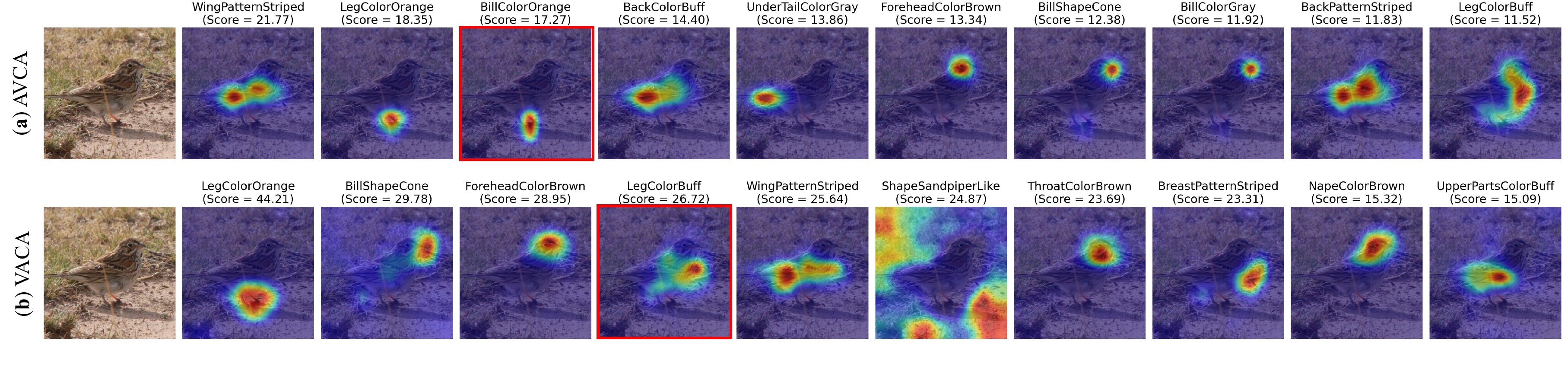}\\\vspace{-4mm}
				\caption{ Visualization of attention maps for the two mutual attention sub-nets (i.e, MSDN++(AVCA) and MSDN++(VACA)).  Results show that our AVCA and VACA subnets can overally learn the accurate visual localizations, but they also learn few of falure cases.}\vspace{-6mm}
				\label{fig:att-2}
			\end{center}
		\end{figure*}

		\noindent\textbf{Conventional Zero-Shot Learning.} 
		We first compare our MSDN++ with the state-of-the-art methods in the CZSL setting. Table \ref{table:sota} shows the results of CZSL on various datasets. Compared to all strong baselines, including embedding-based methods, generative methods, and common space learning methods, our MSDN++ performs the best results of 78.5\%, 67.5\%, and  second-best results of 73.4\% on CUB, SUN, and AWA2, respectively. This indicates that MSDN++ discovers the intrinsic semantic knowledge for effective knowledge transfer from seen classes to unseen ones. Furthermore, our MSDN++ achieves significant performance gains by 2.4\%, 1.7\%, and 3.3\% on CUB, SUN, and AWA2, respectively, over the MSDN (conference version). This should be thanks to the { causal}  visual/attribute leanings that guide MSDN++ to learn causal vision-attribute associations for representing reliable features with good generalization.
		\begin{figure}[htbp]
			\begin{center}
				\includegraphics[width=1.0\linewidth]{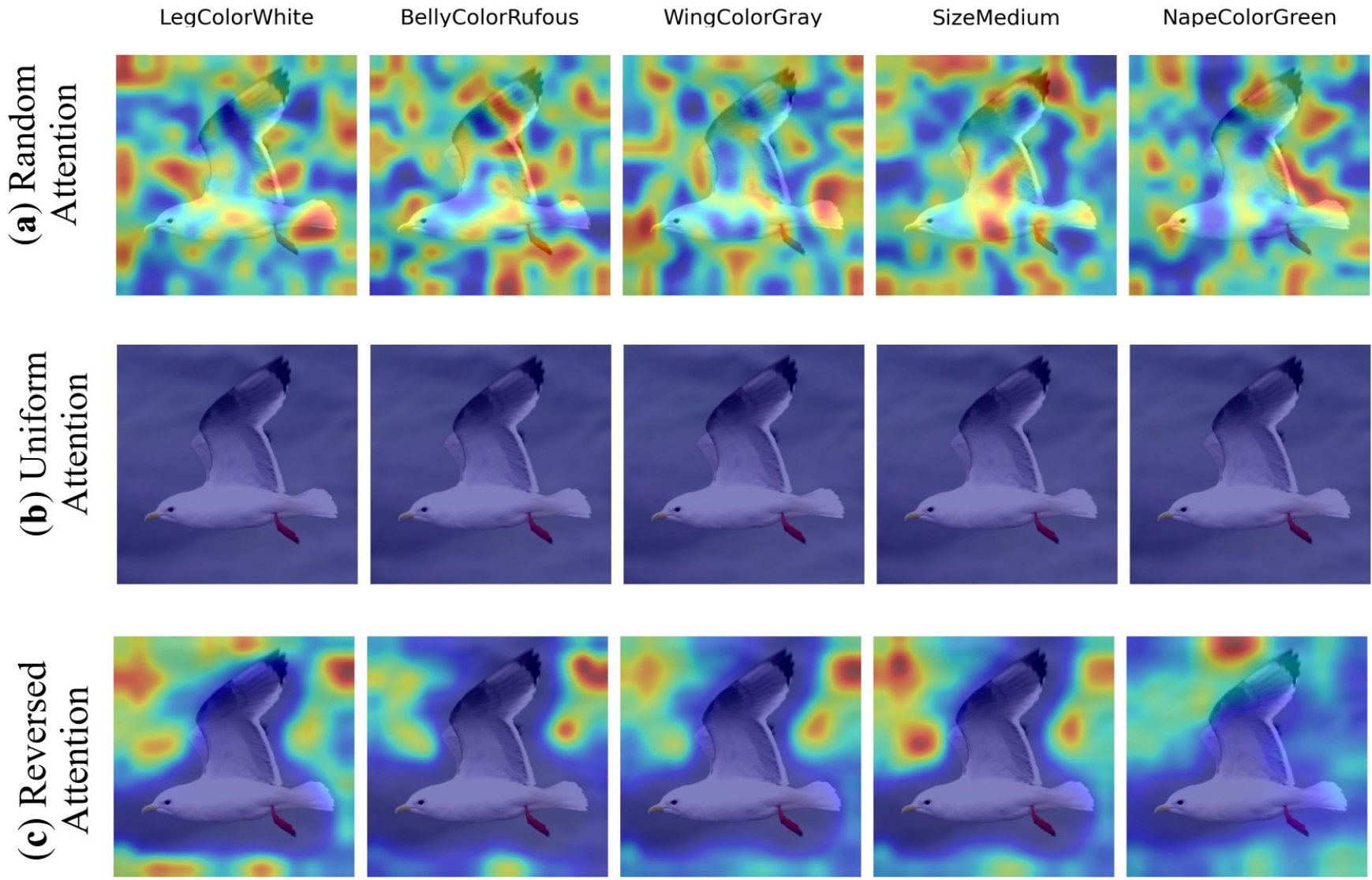} \vspace{-1mm}
				\caption{The attention maps of various { causal}  interventions, including (a) random attention, (b) uniform attention, and (c) reversed attention.}
				\label{fig:counter-att}
			\end{center}
		\end{figure}

		\begin{figure*}[htbp]
			\begin{center}
				\includegraphics[width=0.9\linewidth]{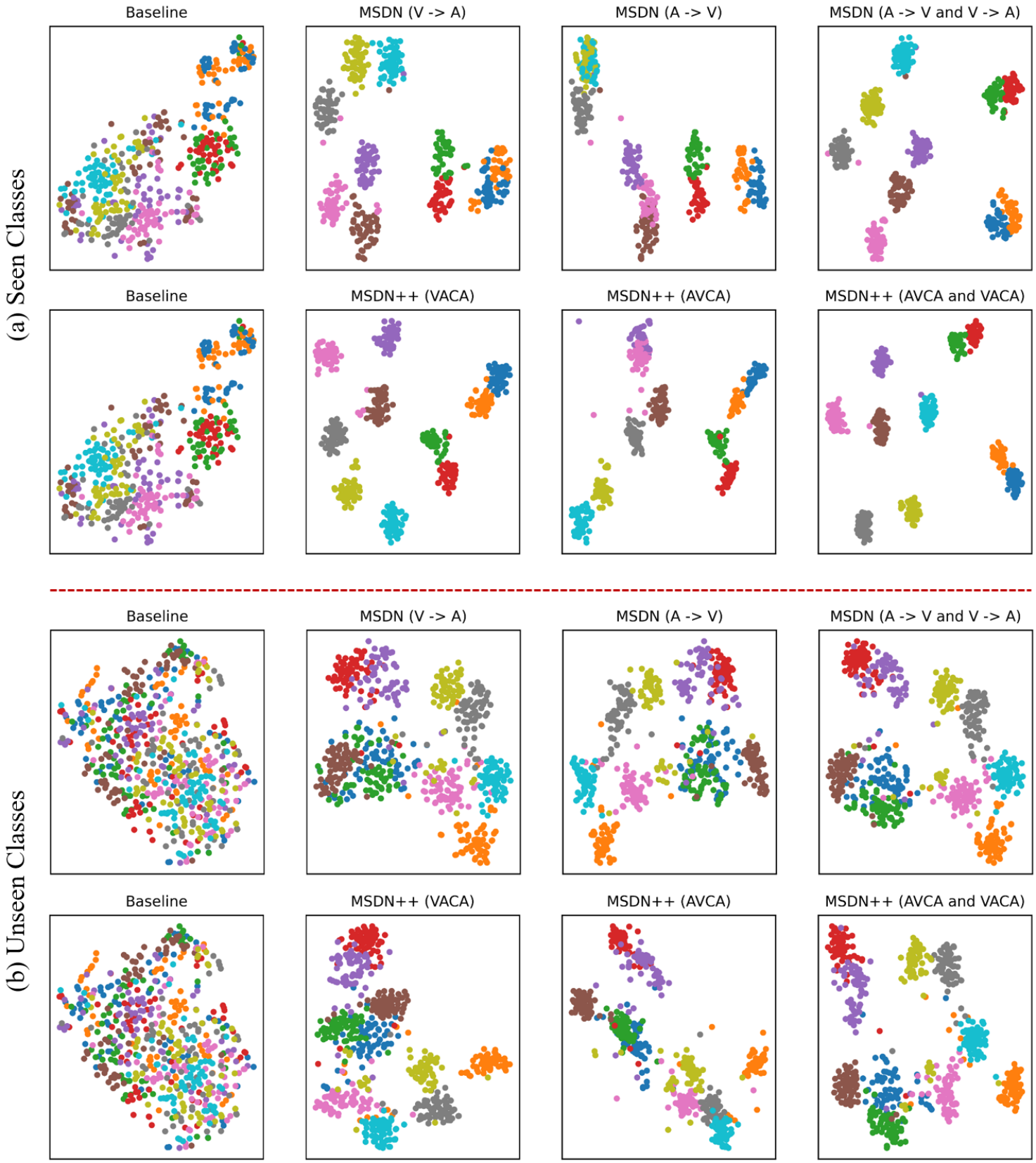}\\
				\caption{t-SNE visualizations of visual features for (a) seen classes and (b) unseen classes, learned by the baseline, MSDN  \citep{Chen2022MSDNMS}, and  MSDN++. The 10 colors denote 10 different seen/unseen classes randomly selected from CUB.  Results show that our MSDN++ learn the intrinsic semantic representations both in seen and unseen classes compared to the baseline. Meanwhile, the each subnet of MSDN++ consistently enhances the intra-class compactness and inter-classes separability for MSDN.}\vspace{-6mm}
				\label{fig:tsne}
			\end{center}
		\end{figure*}
		
		\noindent\textbf{Generalized Zero-Shot Learning.}
		We take MSDN++ to compare with the state-of-the-art methods in the GZSL setting further. Results are shown in Table \ref{table:sota}. We can find:
		
		\noindent\textit{i}) The most state-of-the-art methods achieve good results on seen classes but fail on unseen classes on CUB and AWA2. For example, AREN  \citep{Xie2019AttentiveRE} and LFGAA  \citep{Liu2019AttributeAF} get the accuracies on { unseen/seen classes} are 38.9\%/78.7\% (15.6\%/92.9\%) and 36.2\%/80.9\% (27.0\%/93.4\%) on CUB (AWA2), respectively. Because they simply utilize unidirectional attention in a weakly supervised manner to learn the spurious and limited latent semantic representations, which fails to effectively discover the intrinsic semantic knowledge for knowledge transfer from seen to unseen classes.   In contrast, our MSDN++ employs a mutually { causal}  semantic distillation network to learn intrinsic and more efficient semantic knowledge, which enables the model to generalize well to unseen classes with high seen and unseen accuracies. Accordingly, MSDN++ obtains good results of harmonic mean, \textit{i.e.}, 70.6\%, 42.1\%, and 72.5\% on CUB, SUN, and AWA2, respectively. 
		
		\noindent\textit{ii}) Compared to the large-scale vision-language model based ZSL methods (\textit{e.g.}, CLIP~ \citep{Radford2021LearningTV}, CoOp~ \citep{Zhou2022LearningTP}, CoOp+SHIP  \citep{Wang2023ImprovingZG}), our MSDN++ obtains significant performances gains of $\bm{H}=13.5\%$ at least on CUB. This demonstrates the superiority and potential of our MSDN++ for ZSL. Because the large-scale vision-language based ZSL methods are limited by the domain-specific knowledge, which has a large bias with the large-scale vision-text pairs.

		\noindent\textit{iii}) GCM-CF  \citep{Yue2021CounterfactualZA} is the first method that applied causal inference to the ZSL task. It introduces a causal generation to guide the generative methods to synthesize balanced visual features for unseen classes. Differently, we design { causal} attention learning to enable the embedding-based methods to learn intrinsic and more sufficient semantic knowledge for knowledge transfer from seen to unseen classes. Our MSDN++ achieves significant improvements of harmonic mean by 10.3\% and 5.5\% on CUB and AWA2, respectively. 
		
		{ \noindent\textit{iv}) MSDN++ performs poorer than the generative methods in the GZSL setting because per class only contains 16 training images on SUN, which heavily limits the ZSL models. As such, the data augmentation is very effective for improving the performance of SUN, e.g., the generative methods VS-Boost  \citep{Li2023VSBoostBV}, HSVA  \citep{Chen2021HSVA}. Thus, most of the generative methods perform better than the embedding-based methods on SUN.}
		
		{Additionally, we also conduct experiments on FLO dataset \cite{Nilsback2008AutomatedFC}, results are shown in Table \ref{table:sota-flo}. Results show that our MSDN++ achieves the best result of $\bm{H}=74.5\%$ on FLO. Compared to our conference version (MSDN \cite{Chen2022MSDNMS}), our MSDN++ obtains performance gain of $\bm{H}$ with 4.2\%. These demonstrate the improvements of MSDN++ using causal attention.
		}
		
		\begin{table}[t]
			\small
			\centering
			\caption{ Ablation studies for different components of MSDN++. The \textit{baseline} is the visual feature extracted from CNN backbone with a global average pooling and then mapped into semantic embedding for ZSL.} \label{table:ablation}
			\resizebox{0.49\textwidth}{!}
			{
				\begin{tabular}{l|c|ccc|c|ccc}
					\hline
					\multirow{2}*{Method} &\multicolumn{4}{c|}{CUB} &\multicolumn{4}{c}{AWA2}\\
					\cline{2-5}\cline{6-9}
					&\rm{acc}&\rm{U} & \rm{S} & \rm{H} &\rm{acc}&\rm{U} & \rm{S} & \rm{H}\\
					\hline
					baseline   & 57.4&44.2&55.2&49.1	& 54.8&30.3&30.7&30.5\\
					baseline w/ $\mathcal{L}_{\text{AR}}$& 58.5 &	46.5& 	54.6&	50.2& 	56.9& 	20.6 &	89.7 &	33.5\\ 
					MSDN++(AVCA) w/o  $\mathcal{L}_{\text{distill}}$&76.2 &	68.7 &	69.1& 	68.9& 	71.9& 		65.5 &76.8& 70.7 \\ 
					MSDN++(VACA) w/o $\mathcal{L}_{\text{distill}}$&68.4 & 57.5& 65.6& 		61.3& 	68.0& 		56.5 & 74.6& 	64.3 \\ 
					MSDN++(AVCA) w/ $\mathcal{L}_{\text{distill}}$&77.7 &	69.6& 	70.0& 	69.8& 	73.0& 	66.4& 79.3 &72.3 \\ 
					MSDN++(VACA) w/ $\mathcal{L}_{\text{distill}}$&70.8 &	59.9& 	65.3& 	62.5& 	70.3& 	60.3& 	76.3& 	67.3 \\ 
					MSDN++ w/o AVCA($\mathcal{L}_{\text{causal}}$)&77.5&	67.8& 	71.5 &	69.6& 	72.9& 	66.5& 	77.9& 	71.8 \\ 
					MSDN++ w/o VACA($\mathcal{L}_{\text{causal}}$)&77.9 &	70.6& 	70.2& 	70.4& 	73.3& 	66.9& 	78.4& 	72.2 \\ 
					MSDN++ w/o $\mathcal{L}_{\text{causal}}$&77.0 &69.6& 	69.2& 	69.4& 	72.7& 	66.1& 	78.4& 	71.7 \\ 
					\hline
					MSDN++                                  &\textbf{78.5}&70.8&70.3&\textbf{70.6}&\textbf{73.4}&66.5&79.7&\textbf{72.5}\\
					\hline
				\end{tabular}
			}
		\end{table}
		
			\begin{figure*}[t]
			\begin{center}
				\hspace{-3mm}\includegraphics[width=8.7cm,height=4.1cm]{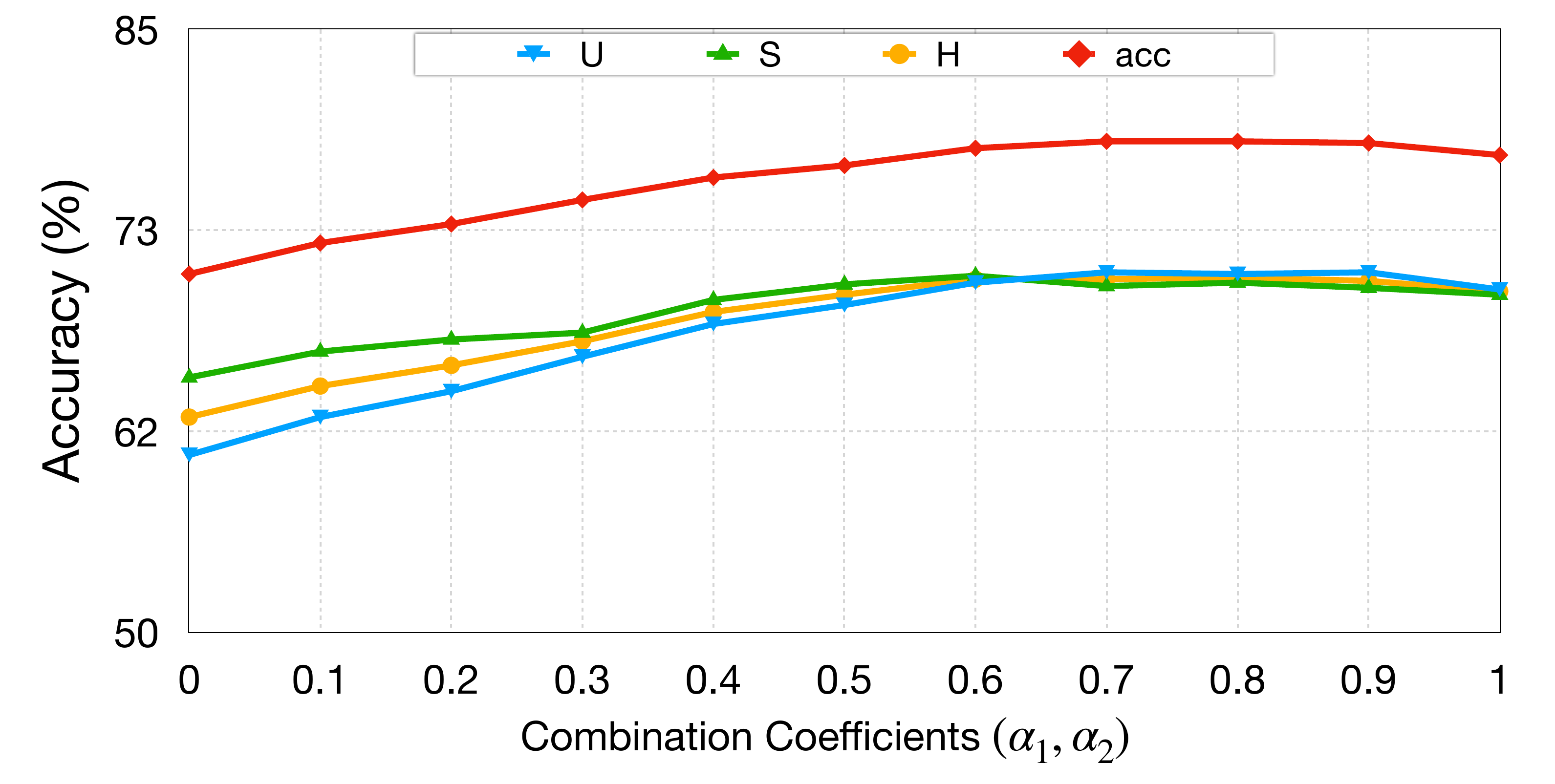}
				\includegraphics[width=8.7cm,height=4.1cm]{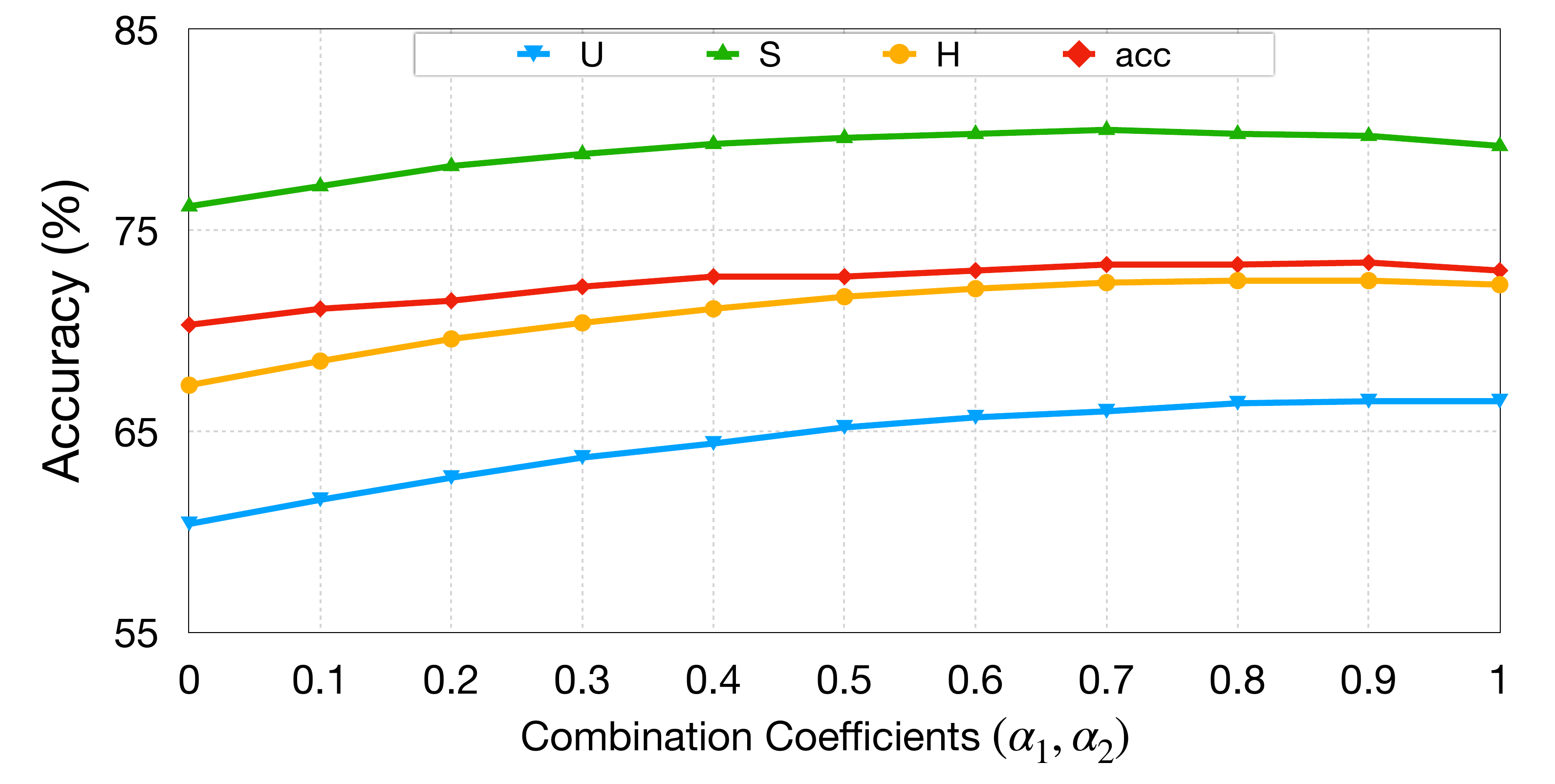}\\ \vspace{-1mm}
				(a) CUB \hspace{8cm} (b) AWA2 \\\vspace{-1mm}
				\caption{The effectiveness of the combination coefficients ($\alpha_{1},\alpha_{2}$) between the AVCA and VACA sub-nets. Results show that MSDN++ performs poorly when $\alpha_{1}/\alpha_{2}$ is set small, and its performances  drop down when $\alpha_{1}/\alpha_{2}$ is set too large. Because  the attribute-based visual features and visual-based attribute features are complementary for discriminative semantic embedding representations.}
				\label{fig:combination}
			\end{center}
		\end{figure*}

		\subsection{Ablation Studies}\label{sec4.2}
		{ We conduct ablation studies to evaluate the effectiveness of various components of MSDN++, including the AVCA attention sub-net (denoted as MSDN(AVCA) w/o $\mathcal{L}_{\text{distill}}$), VACA attention sub-net (denoted as MSDN(VACA) w/o $\mathcal{L}_{\text{distill}}$), semantic distillation learning (\textit{i.e.}, MSDN(VACA) w/ $\mathcal{L}_{\text{distill}}$, MSDN(AVCA) w/ $\mathcal{L}_{\text{distill}}$),  causal visual learning (MSDN++ w/o AVCA($\mathcal{L}_{\text{causal}}$)), causal attribute learning (MSDN++ w/o VACA($\mathcal{L}_{\text{causal}}$)),  and causal learning (\textit{i.e.}, MSDN w/o $\mathcal{L}_{\text{causal}}$). Results are shown in Table \ref{table:ablation}. Compared to the baseline, MSDN++ only employs the single attention sub-net without semantic distillation obtaining significant performance gains. For example,  MSDN(AVCA) w/o $\mathcal{L}_{\text{distill}}$ achieves the gains of $\bm{acc}/\bm{H}$ by 18.8\%/19.8\% and { 17.1}\%/40.2\% on CUB and AWA2, respectively; MSDN++(VACA) w/o $\mathcal{L}_{\text{distill}}$ achieves the $\bm{acc}/\bm{H}$ improvements of 11.0\%/12.2\% and 13.2\%/33.8\% on CUB and AWA2, respectively. Because MSDN++ discovers the intrinsic semantic knowledge using causal learning and visual/attribute-based attribute/visual learning to refine the visual features for effective knowledge transfer. When MSDN++ adopts semantic distillation loss to conduct collaborative learning for knowledge distillation, its results can be further improved, \textit{e.g.},  MSDN(VACA) improves the $\bm{acc}/\bm{H}$ by 2.4\%/1.2\% and 2.3\%/3.0\% on CUB and AWA2, respectively. The causal attribute/visual learning  consistently improve  the performance of MSDN++ by guiding the model to learn causal vision-attribute associations for representing reliable features. Moreover, the two causal learning  can further improve MSDN++ cooperatively. We also find that the model significantly overfits to seen classes without our mutual semantic distillation and causal learning. Our full model ensembles the complementary embeddings learned by the two mutual causal attention sub-nets to represent sufficient semantic knowledge, resulting in further performance gains for MSDN++.}

		\begin{table}[t]
			
			\small
			\centering
			\caption{ Effects of various causal  interventions on CUB, \textit{i.e.}, \textbf{(a) }Random Attention, \textbf{(b)} Uniform Attention, \textbf{(c) }Reversed Attention, and \textbf{(d)} Random Attention+Reversed Attention.} \label{table:counter-att}
			\resizebox{0.49\textwidth}{!}
			{
				\begin{tabular}{l|cccc}
					
					\hline
					\multirow{2}*{Method} &\multicolumn{4}{c}{\textbf{CUB}} \\
					\cline{2-5}
					&\rm{acc}&\rm{U} &\rm{S}& \rm{H}\\
					\hline
					MSDN  \citep{Chen2022MSDNMS}         & 76.1&	68.7 &67.5& 68.1 \\
					\hline
					MSDN++ w/ \textbf{(a)}        & 78.5 &70.8&70.3& 70.6 \\ 
					MSDN++ w/ \textbf{(b) }        & 78.4 &70.8& 70.0&70.4 \\
					MSDN++ w/ \textbf{(c)}        & 78.0 &71.0&	69.9&70.5 \\  
					MSDN++ w/ \textbf{(d) }       & 78.5 &71.8&	69.5&70.6 \\  
					\hline
				\end{tabular}
			}
		\end{table}

		\begin{figure*}[ht]
			\begin{center}
				\hspace{-2.5mm}\includegraphics[width=4.5cm,height=3.35cm]{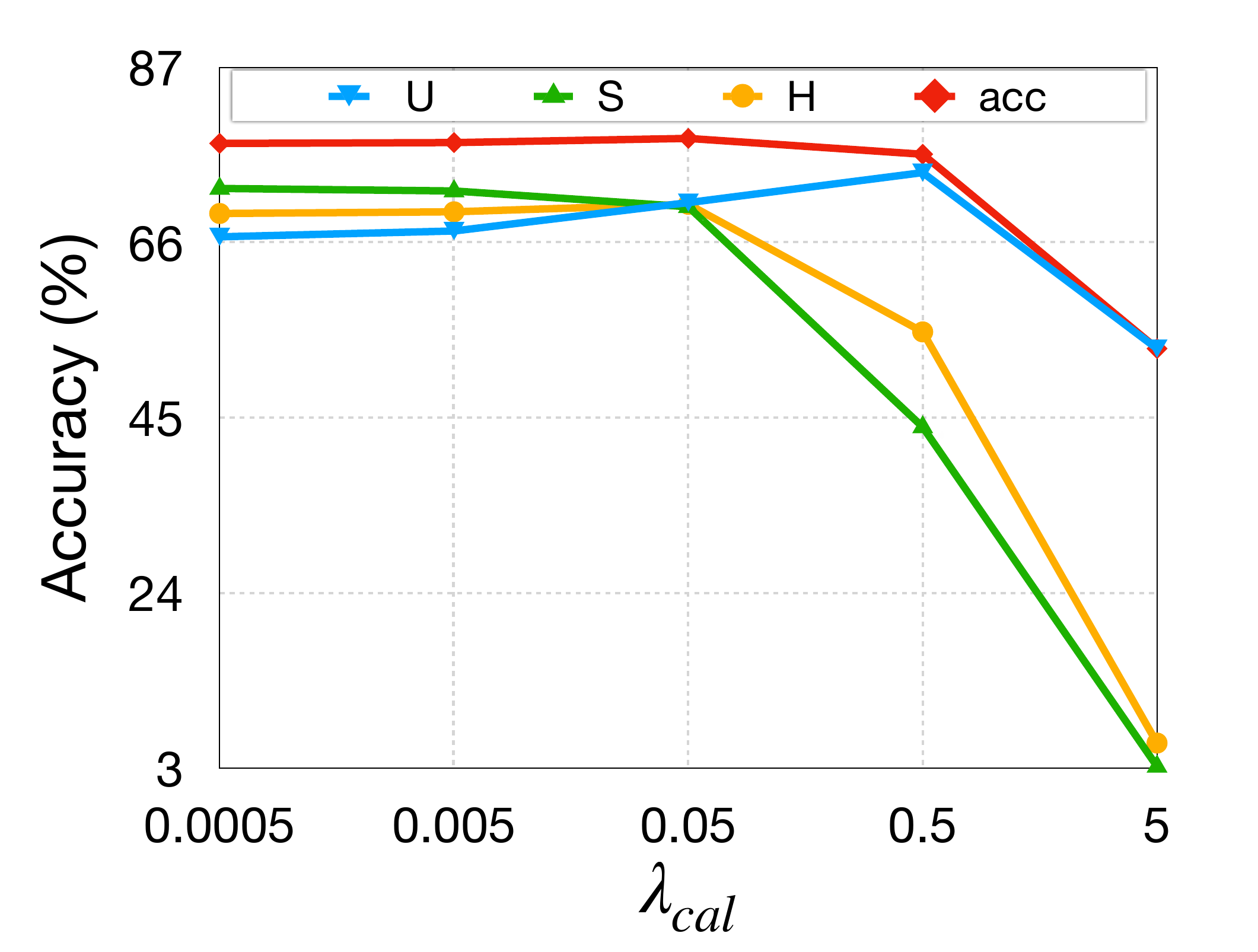}\hspace{-2mm}
				\includegraphics[width=4.5cm,height=3.35cm]{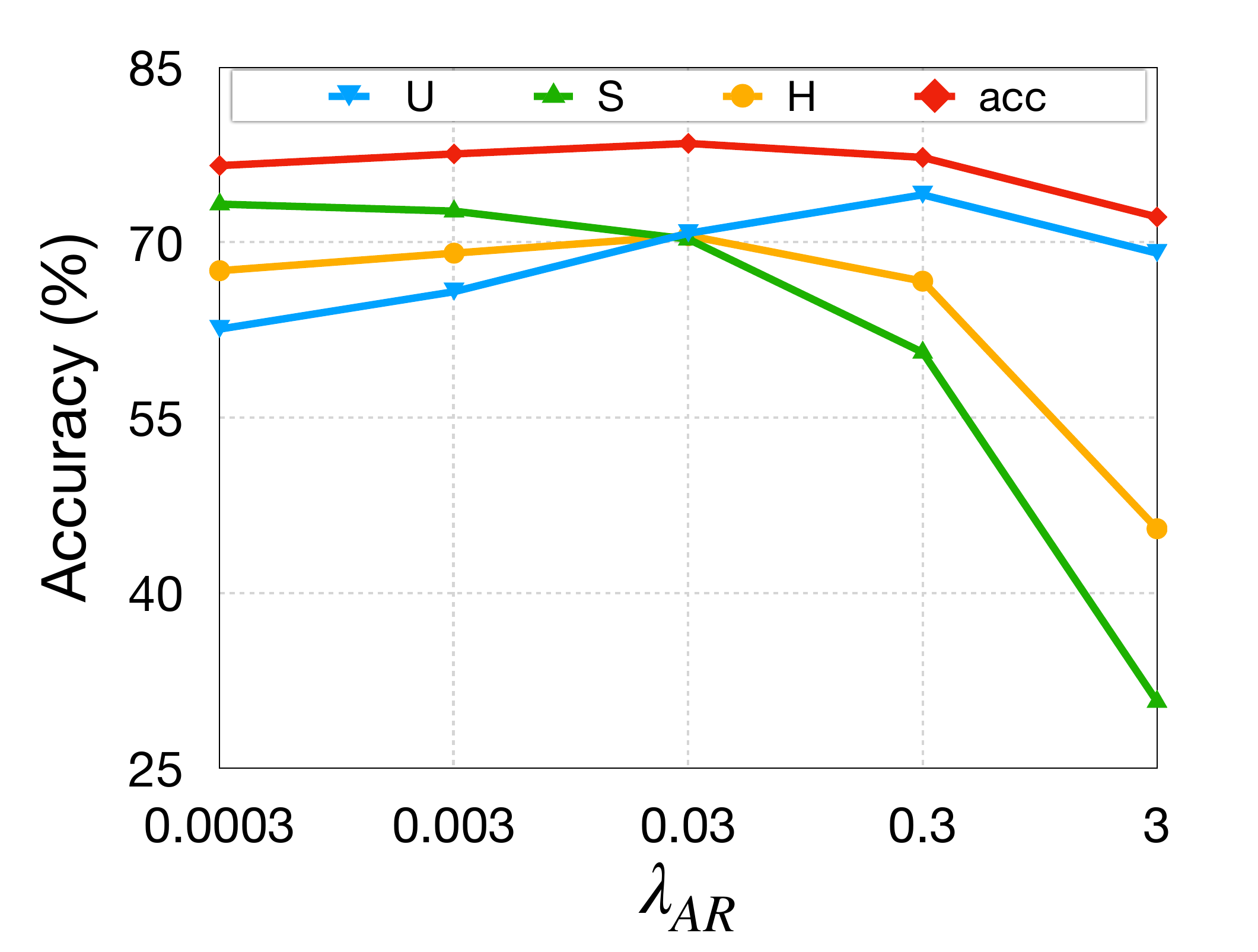}\hspace{-2mm}
				\includegraphics[width=4.5cm,height=3.35cm]{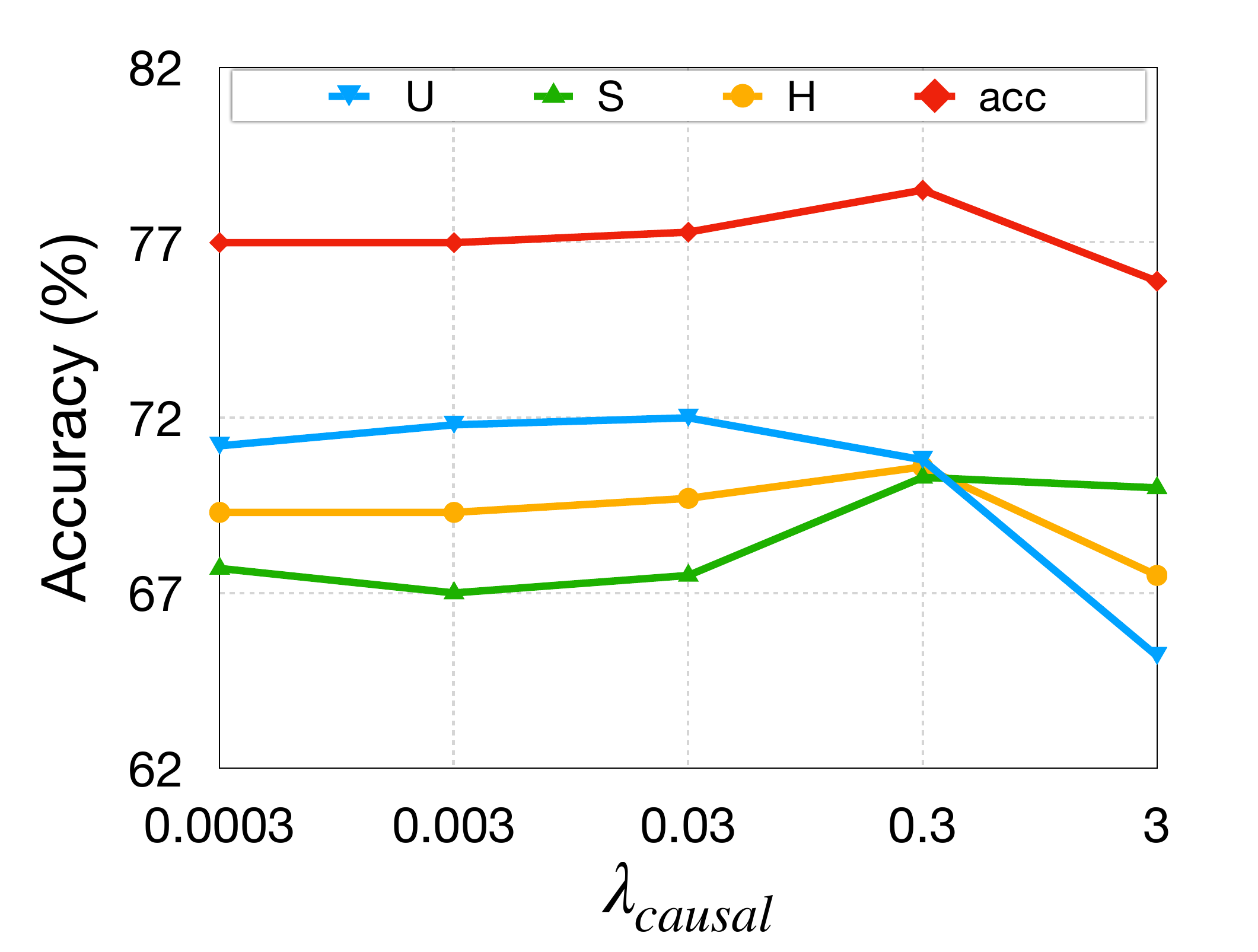}\hspace{-2mm}
				\includegraphics[width=4.5cm,height=3.35cm]{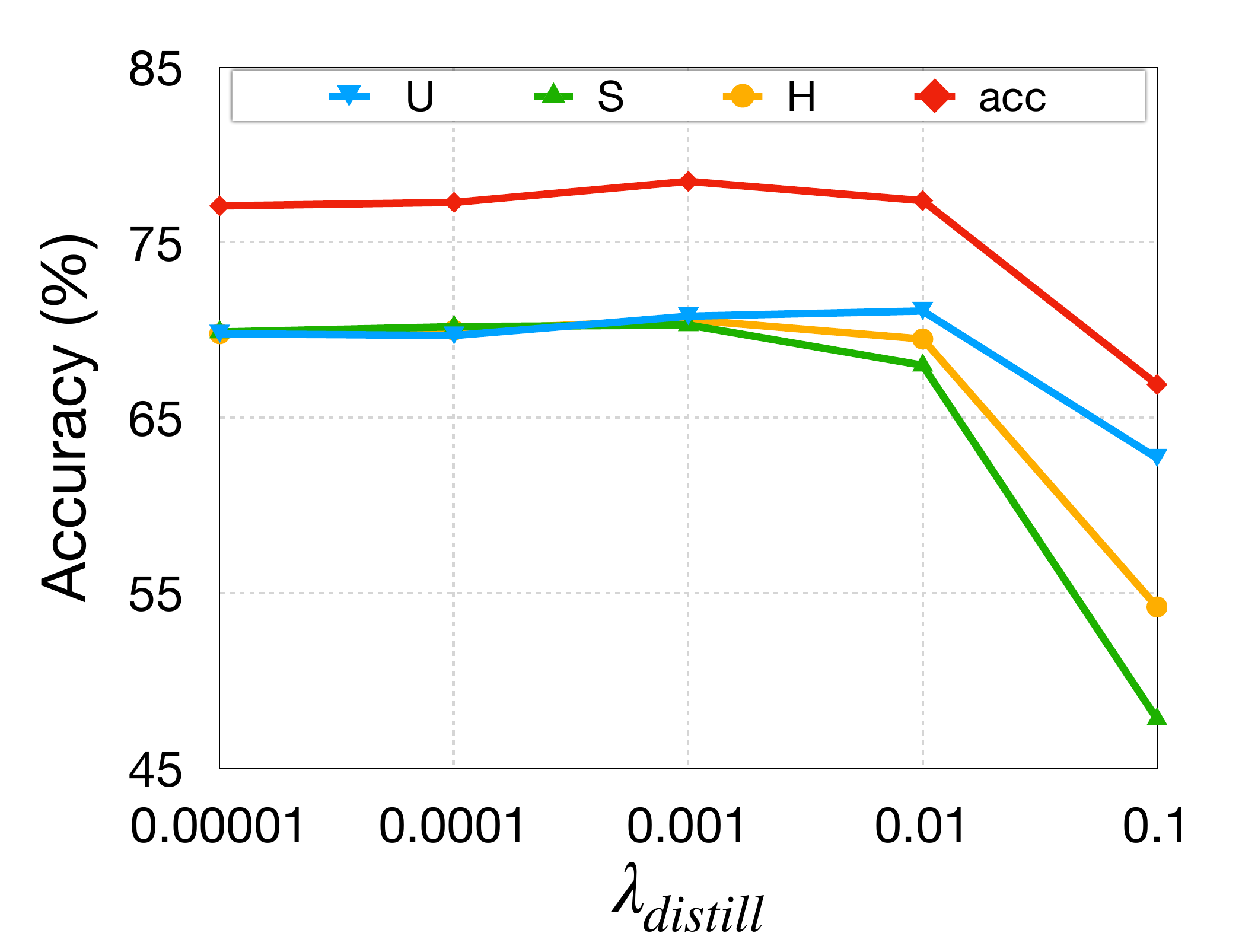}\\ \vspace{-1mm}
				\hspace{-1mm}(a) \hspace{4cm} (b) \hspace{3.8cm} (c) \hspace{3.8cm} (d) \\
				\hspace{-2.5mm}\includegraphics[width=4.5cm,height=3.35cm]{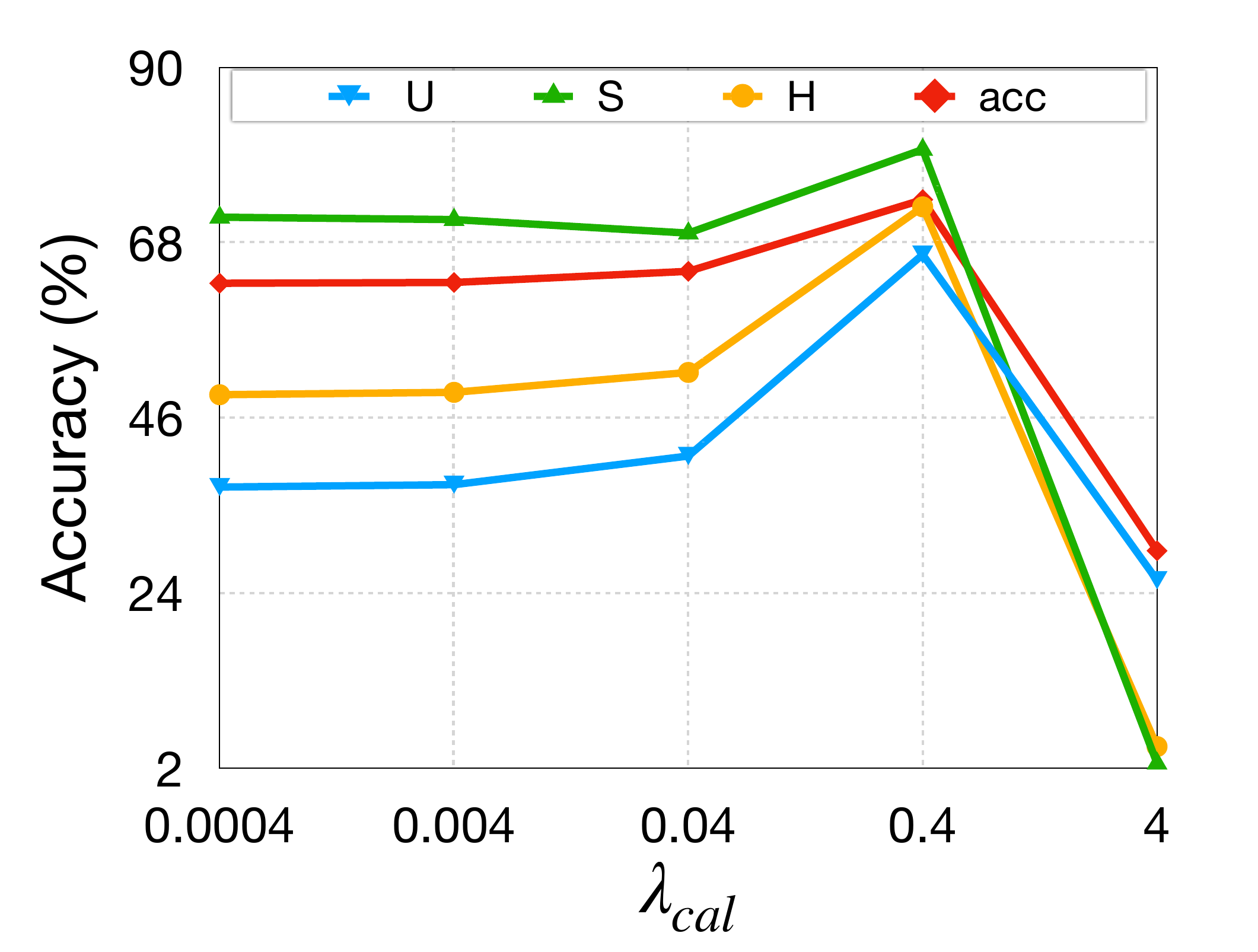}\hspace{-2mm}
				\includegraphics[width=4.5cm,height=3.35cm]{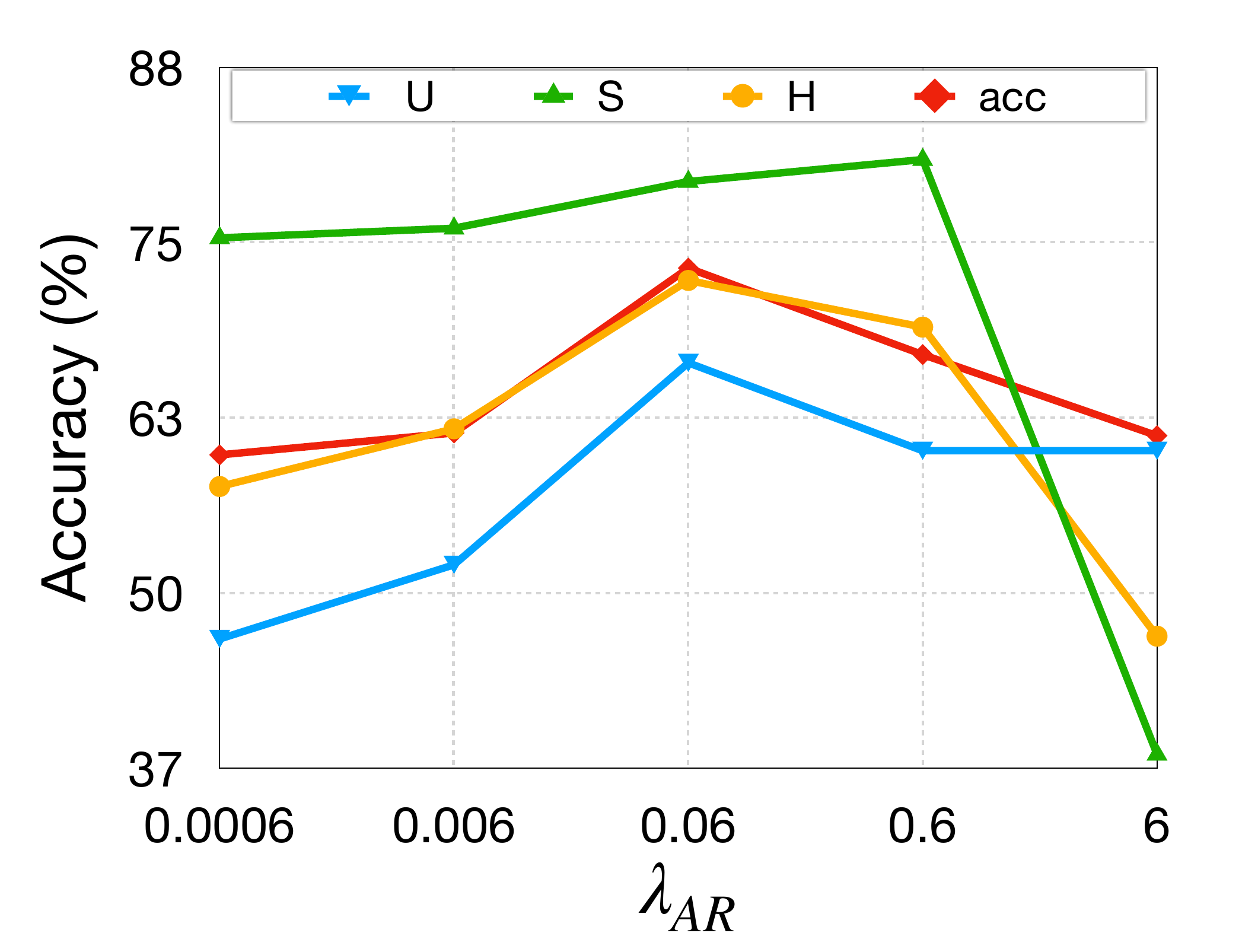}\hspace{-2mm}
				\includegraphics[width=4.5cm,height=3.35cm]{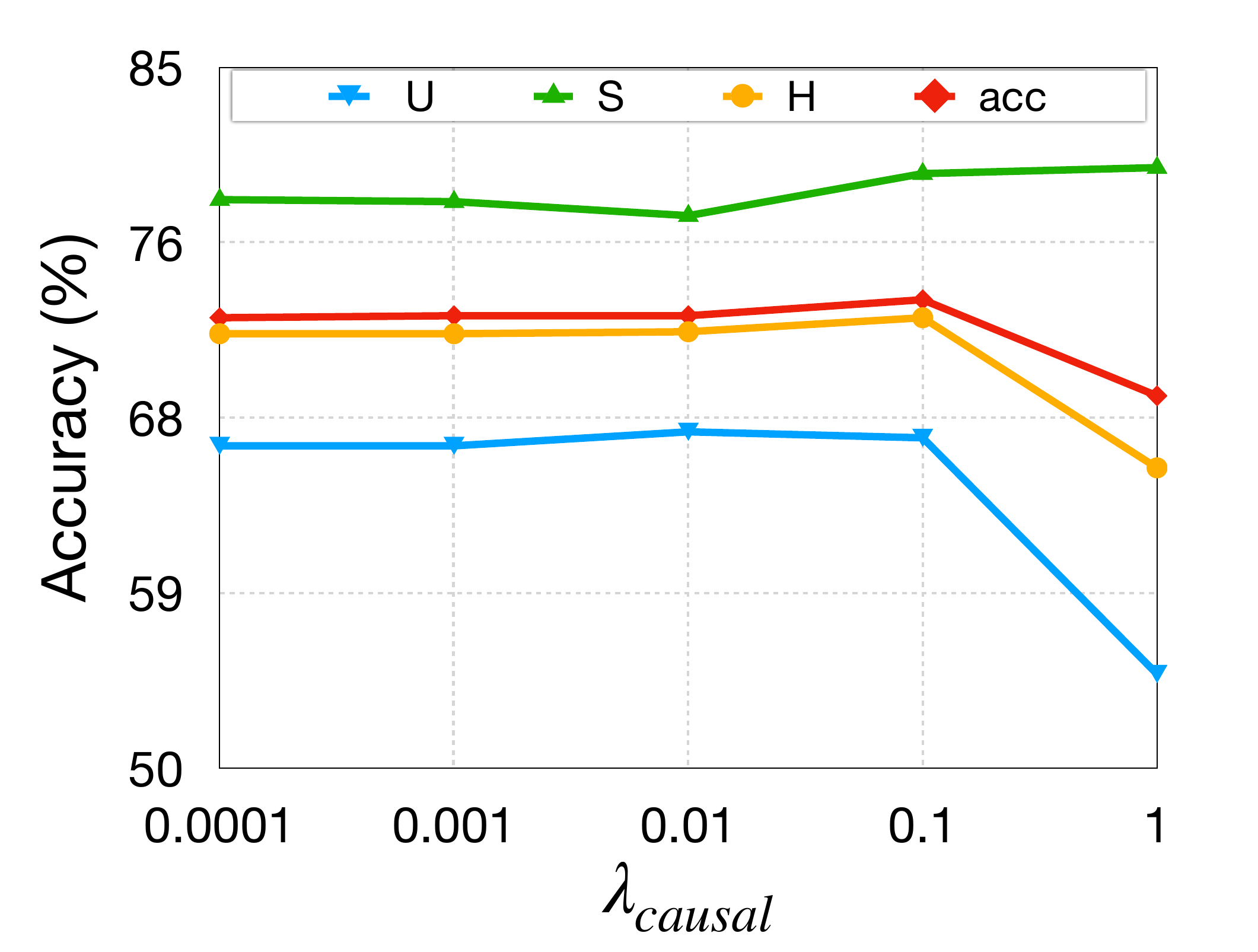}\hspace{-2mm}
				\includegraphics[width=4.5cm,height=3.35cm]{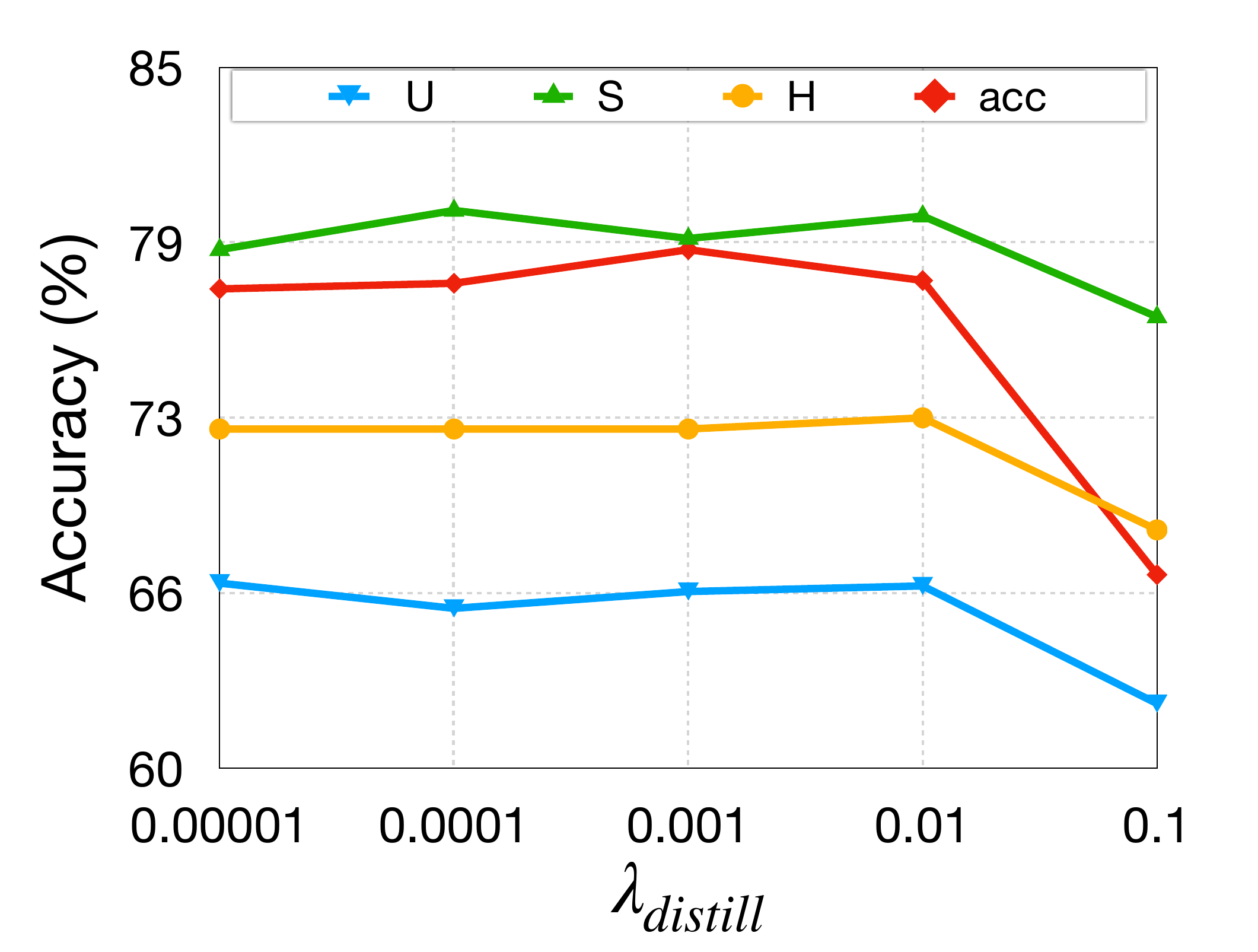}\\ \vspace{-1mm}
				\hspace{-1mm}{ (e) }\hspace{4cm} { (f) }\hspace{3.8cm} (g) \hspace{3.8cm} (h)
				\caption{The effects of $\lambda_{\text{cal}}$, $\lambda_{\text{AR}}$, $\lambda_{\text{causal}}$, and $\lambda_{\text{distill}}$ on CUB (top-row) and AWA2 (bottom-row). Results show that MSDN++ is robust when the loss weights are set to small and the performance will drop rapidly when loss weights are set to too large. Because the large loss weights will hamper the balance of various losses.}
				\label{fig:loss_cal}
			\end{center}
		\end{figure*}
		
		\subsection{Qualitative Results}\label{sec4.3}
		\noindent\textbf{Visualization of Attention Maps.}
		To intuitively show the effectiveness of our MSDN++ for learning more intrinsic semantic knowledge than MSDN  \citep{Chen2022MSDNMS}, we visualize the top-10 attention maps learned by the two methods. As shown in Fig. \ref{fig:att}, although MSDN can localize some attributes in the image correctly for semantic knowledge representations, a few of the attributes are localized wrongly. For example, the MSDN localizes the attribute “leg color brown” of \textit{Acadian Flycatcher} on the head in the image. This is because MSDN learns the wrong associations between visual and attribute features for representing spurious semantic knowledge. In contrast, our MSDN++ learns the important attribute localization accurately for intrinsic semantic knowledge representations using { causal}  attribute/visual learning, which encourages MSDN++ to discover the causal vision-attribute associations for representing reliable features. { As shown in Fig. \ref{fig:att-2}, the two sub-nets of MSDN++ can similarly learn the most important semantic representation, which is beneficial from mutual learning for semantic distillation. Furthermore, the two attention sub-nets also learn the complementary attribute feature localization for each other. } As such, our MSDN++ achieves significant performance gains over MSDN.
		
		{Although our AVCA and VACA subsets can overally learn the accurate visual localizations, they also learn few of failure cases. For example, the attribute “\textit{Bill Color Orange}” and the attribute “\textit{Leg Color Buff}” are wrongly localized by AVCA and VACA, respectively.  This may be caused by the semantic  ambiguity, which denotes that the same attribute may represented as various visual appearances.  We believe this is an open challenge for future works.}

		\noindent\textbf{t-SNE Visualizations.}
		As shown in Fig. \ref{fig:tsne}, we also present the t-SNE visualization  \citep{Maaten2008VisualizingDU} of visual features for seen and unseen classes on CUB, learned by the baseline, MSDN  \citep{Chen2022MSDNMS}, and MSDN++. Compared to the baseline, our models learn the intrinsic semantic representations both in seen and unseen classes. For each sub-net, MSDN++ consistently enhances the intra-classes compactness and inter-classes separability for MSDN, and thus the fused features are refined further. This should be thanks to our { causal}  visual/attributes learning in the AVCA/VACA subnets, which encourage MSDN++ to learn causal vision-attribute associations for representing reliable features with good generalization.  As such, our MSDN++ achieves significant improvement over MSDN and baseline.

		\subsection{Effects of Various { Causal}  Intervention}\label{sec4.5}
		We analyze three different strategies to conduct a { causal}  intervention, including random attention, uniform attention and reversed attention, which will generate various { causal}  attention maps as shown in Fig. \ref{fig:counter-att}. Results show that the { causal}  attention maps focus on the wrong visual regions, which are not relevant to their corresponding attributes. This causal can measure the effectiveness of our attribute-based visual learning and visual-based attribute learning and enable them to learn the intrinsic semantic knowledge between visual and attribute features with causality. The results in Table \ref{table:counter-att} show that the various { causal}  attentions consistently improve the performance of MSDN. Because our { causal}  attention provides a significant signal to supervise MSDN++. Furthermore, we find that all the causal interventions get similar performance gains, which indicates our { causal}  visual/attribute learning is robust for improving the causality of features. 
		

		\subsection{Hyperparameter Analysis}\label{sec4.4}
		\noindent\textbf{Effects of Combination Coefficients.}
		we provide experiments to determine the effectiveness of the combination coefficients ($\alpha_{1},\alpha_{2}$) between AVCA and VACA sub-nets.  As shown in Fig. \ref{fig:combination}, MSDN++ performs poorly when $\alpha_{1}/\alpha_{2}$ is set small, because the AVCA is the main sub-net to support the final classification of MSDN++. Since the attribute-based visual features and visual-based attribute features are complementary for discriminative semantic embedding representations, the performances of MSDN++ drop down when $\alpha_{1}/\alpha_{2}$ is set too large (\textit{e.g.}, $(\alpha_{1},\alpha_{2})=(0.9,0.1)$). { According to Fig. \ref{fig:combination},   we set $(\alpha_{1},\alpha_{2})$ to be (0.8,0.2) for CUB and AWA2 based on the performances of $\bm{H}$ and $acc$}. Notably, $(\alpha_{1},\alpha_{2})=(1.0,0.0)$ and $(\alpha_{1},\alpha_{2})=(0.0,1.0)$ denotes the MSDN++ without information fusion from AVCA and VACA during inference.

		\noindent\textbf{Effects of Loss Weights.} We empirically study how to set the related loss weights of MSDN++: $\lambda_{\text{cal}}$,  $\lambda_{\text{AR}}$,  $\lambda_{\text{causal}}$, and $\lambda_{\text{dstill}}$, which control the self-calibration term, attribute regression loss, causal loss, and semantic distillation loss, respectively. Results are shown in Fig. \ref{fig:loss_cal}. Results show that MSDN++ is robust when the loss weights are set to small and the performance will drop rapidly when loss weights are set to too large. Because the large loss weights will hamper the balance of various losses. According to the results in Fig. \ref{fig:loss_cal}, we set the loss weights $\{\lambda_{\text{cal}},\lambda_{\text{AR}},\lambda_{\text{causal}},\lambda_{\text{distill}}\}$ to $\{0.05,0.03,0.3,0.001\}$ and $\{0.4,0.06,0.1,0.01\}$ for CUB and AWA2, respectively. Notably, the loss weights $\{\lambda_{\text{cal}},\lambda_{\text{AR}},\lambda_{\text{causal}},\lambda_{\text{distill}}\}$ are to $\{0.0001,0.01,0.0005,0.05\}$ for SUN.  {Furthermore, although the performance curve somehow unstale with the variant values for the hyper-parameters, MSDN++ performs consistently on various datasets when using same hyper-parameters. This means that our MSDN++ is roubust on various datasets. Accordingly, our method can be easily applied into new domains.}

		\section{Conclusion}\label{sec5}
		
		This paper introduces a mutually { causal}  distillation network for ZSL, termed MSDN++, which consists of an AVCA and a VACA sub-nets. AVCA learns the attribute-based visual features with attribute-based/{ causal}  visual learnings, while VACA learns the visual-based attribute features via visual-based/{ causal}  attribute learnings. The { causal}  attentions encourage the two sub-nets to discover { causal}  vision-attribute associations for representing reliable features with good generalization. Furthermore, we further introduce a semantic distillation loss, which promotes the two sub-nets to learn collaboratively and teach each other. Finally, we fuse the complementary features of the two sub-nets to make full use of all important knowledge during inference. As such, MSDN++ effectively explores the intrinsic and more sufficient semantic knowledge for desirable knowledge transfer in ZSL. The quantitative and qualitative results on three popular benchmarks (\textit{i.e.}, CUB, SUN, and AWA2) demonstrate the superiority and potential of our MSDN++.

	\section*{Data Availability Statements}\label{sec7}
	The data used in this manuscript includes three benchmark datasets, i.e., CUB,\footnote{\url{ https://www.vision.caltech.edu/datasets/cub_200_2011/}} SUN,\footnote{\url{ https://cs.brown.edu/~gmpatter/sunattributes.html}} and AWA2,\footnote{\url{https://cvml.ist.ac.at/AwA2/}} which are publicly opened benchmarks.

	
	\bibliographystyle{spbasic}      
	\bibliography{MSDN-V2}   

\end{document}